\documentclass[11pt]{article}
\usepackage{enumerate}
\usepackage{pdfsync}
\usepackage[OT1]{fontenc}

\usepackage[usenames]{color}
\usepackage{smile}
\usepackage[colorlinks,
            linkcolor=red,
            anchorcolor=blue,
            citecolor=blue
            ]{hyperref}
\usepackage{fullpage}
\usepackage[protrusion=true,expansion=true]{microtype}
\usepackage{pbox}
\usepackage{setspace}
\usepackage{tabularx}
\usepackage{float}
\usepackage{wrapfig,lipsum}
\usepackage{enumitem}
\usepackage{graphicx}
\usepackage{subfigure}
\usepackage{pgfplots}
\usetikzlibrary{arrows,shapes,snakes,automata,backgrounds,petri}
\usepackage{booktabs} 
\usepackage{mathtools}

\allowdisplaybreaks
\usepackage{colortbl}
\definecolor{LightCyan}{rgb}{0.8, 0.9, 1}
\definecolor{LightGray}{gray}{0.9}

\usepackage{xcolor}

\ifdefined\final
\usepackage[disable]{todonotes}
\else
\usepackage[textsize=tiny]{todonotes}
\fi
\makeatletter
\newcommand*{\rom}[1]{\expandafter\@slowromancap\romannumeral #1@}
\makeatother

\providecommand{\Pg}{\mathcal{P}}
\providecommand{\kt}{\tilde k}
\providecommand{\gh}{\hat g}
\providecommand{\ghb}{\hat{\gb}}
\renewcommand{\brho}{\bm{\rho}}
\providecommand{\lmax}{\lambda_{\max}}
\renewcommand{\norm}[1]{\left\|#1\right\|}
\newcommand{\lstep}[1]{\overset{\text{(#1)}}{\le}}
\newcommand{\gstep}[1]{\overset{\text{(#1)}}{\ge}}

\newcommand{\estep}[1]{\overset{\text{(#1)}}{=}}

\newcommand{\sstep}[1]{\overset{\text{(#1)}}{\succeq}}

\title{\huge Nearly Minimax Optimal Regret for Adversarial Linear Bandits with Strong Corruptions}
\author
{
   Qiwei Di\thanks{Department of Computer Science, University of California, Los Angeles, CA 90095, USA; e-mail: {\tt qiwei2000@cs.ucla.edu}} 
	~~~and~~~
	Quanquan Gu\thanks{Department of Computer Science, University of California, Los Angeles, CA 90095, USA; e-mail: {\tt qgu@cs.ucla.edu}}
}
\begin{document}
    \date{}
    \maketitle

\begin{abstract}
We study adversarial linear bandits with strong corruption, in which an adaptive adversary chooses the reward vectors and a strong adversary corrupts each observed reward after seeing the learner's action, subject to a known total corruption budget $C$. Strong corruption breaks the importance-weighted reward estimates used by adversarial bandit algorithms: by corrupting only the rounds in which a rarely sampled action is drawn, the adversary amplifies each unit of corruption by the inverse of its sampling probability. The best known bound for this problem, $\tilde O(\sqrt{d^3T} + dC)$ by \citet{liu2024corruption}, is a factor $\sqrt{d}$ above the corruption-free rate in its leading term, and whether $\tilde O(d\sqrt{T} + dC)$ regret is achievable was left open. We propose CRW-FTRL, which runs follow-the-regularized-leader with a radius-weighted entropy regularizer and estimates rewards with a normalized kernel estimator whose sensitivity to corruption is governed by per-action radii instead of inverse sampling probabilities. The radii are nondecreasing and are chosen jointly with the sampling distribution through a monotone log-determinant barrier problem, so that a negative linear term in the regularizer cancels both the corruption bias and the estimation error of every comparator. For $K$ actions in $\RR^d$, CRW-FTRL achieves $\tilde O(\sqrt{dT\log K} + dC)$ regret with high probability against an adaptive adversary, which matches the best known corruption-free rate and the $\Omega(dC)$ lower bound up to logarithmic factors. Applied to a discretization of a general compact action set, it yields $\tilde O(d\sqrt{T} + dC)$ regret, which settles the statistical rate of corrupted adversarial linear bandits under strong corruption.
\end{abstract}

\section{Introduction}\label{sec:intro}
Adversarial linear bandits \citep{awerbuch2004adaptive,mcmahan2004online,dani2007price,abernethy2008competing,bubeck2012towards} are a basic model of sequential decision making with structured actions and nonstationary rewards. In each round, the learner picks an action $\ab \in \cA \subset \RR^d$, an adversary picks a reward vector $\btheta_t \in \RR^d$, and the learner observes only the reward $\ab^\top \btheta_t$ of the action it played, possibly with noise. The regret of this problem is well understood. For a finite action set of size $K$, exponential weights with a suitable exploration distribution achieves $O(\sqrt{dT\log K})$ regret \citep{bubeck2012towards}, which is minimax optimal up to logarithmic factors on the hypercube \citep{dani2007price}, and a rate of the same order holds with high probability against an adaptive adversary \citep{zimmert2022return}. These guarantees all assume, however, that the observed reward equals the linear reward of the chosen action plus zero-mean noise.

In many applications, this assumption fails because the feedback itself is manipulated. Click fraud in online advertising, fake ratings in recommender systems, and malicious users who poison the feedback of an online learner all inject corruptions into the observed rewards, and even a small corruption budget suffices to mislead standard bandit algorithms \citep{lykouris2018stochastic,jun2018adversarial,liu2019data,garcelon2020adversarial}. Corruption-robust bandit algorithms address this issue by allowing an adversary to perturb the observed rewards subject to a total budget $C$, and they aim for regret that degrades gracefully with $C$. For stochastic linear bandits, this goal is largely achieved: \citet{he2022nearly} obtain $\tilde O(d\sqrt{T} + dC)$ regret, which matches the $\Omega(d\sqrt{T} + dC)$ lower bound \citep{lattimore2020bandit,bogunovic2021stochastic} up to logarithmic factors. When the rewards are themselves adversarial, the corruption cannot be absorbed into the reward vectors $\btheta_t$, because it may depend on the realized action and is not linear in the action \citep{liu2024corruption}. Corrupted adversarial linear bandits are therefore a genuinely different problem, and they were studied only recently.

\citet{liu2024corruption} initiated the study of corrupted adversarial linear bandits and distinguished two types of adversaries. A \emph{weak} adversary fixes the corruption of every action before the learner draws its action, whereas a \emph{strong} adversary chooses the corruption after observing the drawn action. Against a weak adversary, \citet{liu2024corruption} obtain $\tilde O(d\sqrt{T} + \sqrt{d}C_\infty)$ regret, matching the stochastic case. Against a strong adversary, their bound is $\tilde O(\sqrt{d^3T} + dC)$, whose leading term is a factor $\sqrt{d}$ larger than the corruption-free rate, and they leave open whether $d\sqrt{T} + dC$ regret is achievable \citep[Section~7]{liu2024corruption}. The difficulty lies in the reward estimator. Adversarial bandit algorithms rely on importance-weighted estimates of the rewards, and the weight placed on the observed reward can be as large as $1/p_{t,I_t}$ when the drawn action $I_t$ has small sampling probability $p_{t,I_t}$. A weak adversary cannot exploit this, because the large weight is drawn with probability $p_{t,I_t}$ and cancels in expectation. A strong adversary, by contrast, can stay silent in most rounds and spend its budget only in rounds where a rarely sampled action is drawn, so that each unit of corruption shifts the reward estimates by $1/p_{t,I_t}$ instead of $O(1)$. \citet{liu2024corruption} avoid this problem by running continuous exponential weights over a lifted action space, whose concentrated sampling keeps the weights bounded by $\mathrm{poly}(d)$, and their analysis of this approach incurs the additional $\sqrt{d}$ factor in the leading term.

In this paper, we resolve this open question in the affirmative. We study adversarial linear bandits with $K$ actions, where an adaptive adversary chooses the reward vectors and a strong adversary chooses the corruptions after observing both the drawn action and its clean reward, subject to a known budget $C$. Our contributions are summarized as follows.
\begin{itemize}[leftmargin=*]
  \item We propose CRW-FTRL, an FTRL algorithm that combines a radius-weighted entropy regularizer with normalized kernel reward estimates. In each round, the algorithm assigns each action a nondecreasing radius via a monotone log-determinant barrier problem and constructs a reweighted, normalized kernel reward estimator. Controlling the estimator's sensitivity through these radii rather than inverse sampling probabilities prevents a strong adversary from amplifying its corruption on rarely sampled actions, while the radius-weighted entropy regularizer with a linear bonus offsets both the cumulative corruption bias and the estimation variance.
  \item We prove that, with probability at least $1-\delta$, CRW-FTRL achieves realized regret $O\big(dC\log(dT) + \sqrt{dT\log(dT)\log(KT/\delta)}\big) = \tilde O(\sqrt{dT\log K} + dC)$ against an adaptive adversary with strong corruption (Theorem~\ref{thm:main} and Corollary~\ref{cor:rate}). The corruption-free term matches the best known high-probability rate $\tilde O(\sqrt{dT\log K})$ for finite action sets \citep{bubeck2012towards,zimmert2022return}, and the corruption term matches the $\Omega(dC)$ lower bound of \citet{bogunovic2021stochastic} (which holds already for $O(d)$ actions; adjoining the zero action to their $d$ standard-basis arms gives $K = d + 1$ actions with full affine span in $\RR^d$), both up to logarithmic factors.
  \item Because $K$ enters our bound only through $\log K$, running CRW-FTRL on a fine discretization of a compact action set yields $\tilde O(d\sqrt{T} + dC)$ regret for general action sets with bounded reward vectors (Remark~\ref{rem:main-discussion}). This answers the question of \citet{liu2024corruption} on the statistical rate and improves the leading term of their bound from $\sqrt{d^3T}$ to $d\sqrt{T}$. Together with their result for weak adversaries, it shows that corrupted adversarial linear bandits attain the same minimax rates as corrupted stochastic linear bandits under both corruption models. Since the discretization has exponentially many actions in $d$, this reduction is not computationally efficient, and achieving the same rate with a computationally efficient algorithm remains open.
\end{itemize}

\noindent\textbf{Notation.}
For nonnegative sequences $\{x_n\}$ and $\{y_n\}$, we write $x_n = O(y_n)$ if $\limsup_{n\to\infty}{x_n}/{y_n} < \infty$, $y_n = \Omega(x_n)$ if $x_n = O(y_n)$, and $y_n = \Theta(x_n)$ if $x_n = O(y_n)$ and $x_n = \Omega(y_n)$. We further employ $\tilde{O}(\cdot), \tilde{\Omega}(\cdot)$, and $\tilde{\Theta}(\cdot)$ to hide polylogarithmic factors.
We denote $[N] \coloneqq \{1, \cdots, N\}$ for any positive integer $N$, and write $\Delta_K \coloneqq \{\pb \in \RR^K : p_i \ge 0,\ \sum_{i=1}^K p_i = 1\}$ for the probability simplex over $[K]$.
Boldfaced lowercase and uppercase letters are reserved for vectors and matrices, respectively. We denote by $\eb_i$ the $i$-th standard basis vector, by $\mathbf 0$ and $\mathbf 1$ the all-zeros and all-ones vectors, by $\norm{\cdot}$ the Euclidean norm, and by $\inner{\xb}{\yb} \coloneqq \xb^\top \yb$ the standard inner product for vectors $\xb, \yb$ of the same dimension. For symmetric matrices $\Ab$ and $\Bb$, we write $\Ab \succeq \Bb$ (resp.\ $\Ab \succ \Bb$) if $\Ab - \Bb$ is positive semidefinite (resp.\ positive definite), and use $\lambda_{\max}(\cdot)$, $\tr(\cdot)$, and $\det(\cdot)$ to denote the largest eigenvalue, trace, and determinant.

\section{Related Work}\label{sec:related}

\paragraph{Adversarial Linear Bandits.}
Adversarial linear bandits were introduced by \citet{awerbuch2004adaptive} and \citet{mcmahan2004online}. \citet{dani2007price} proposed GeometricHedge, which runs exponential weights over the action set and explores with a barycentric spanner, and proved an $\Omega(d\sqrt{T})$ lower bound on the hypercube with i.i.d.\ losses. \citet{abernethy2008competing} gave a computationally efficient algorithm with $O(d\sqrt{\vartheta T\log T})$ regret by running FTRL with a $\vartheta$-self-concordant barrier (yielding $\tilde O(d\sqrt{T})$ when $\vartheta = O(1)$ and $\tilde O(d^{3/2}\sqrt{T})$ for a general $\vartheta = O(d)$ barrier). \citet{cesa2012combinatorial} and \citet{bubeck2012towards} refined the exploration distribution of exponential weights, and \citet{bubeck2012towards} obtained $O(\sqrt{dT\log K})$ regret for $K$ actions by exploring according to John's theorem. High-probability bounds against adaptive adversaries are more delicate, and existing approaches control the variance of importance weighting either through biased reward estimates and bonuses \citep{bartlett2008high,zimmert2022return} or by combining unbiased estimators with an increasing learning rate schedule \citep{lee2020bias}. \citet{zimmert2022return} obtained almost minimax optimal high-probability bounds, using exponential weights for finite action sets and, for general action sets, FTRL with a log-determinant potential over lifted covariance matrices of the action distribution. In the presence of corruptions, \citet{liu2024corruption} extended this lifted log-determinant FTRL approach to weak adversaries and used continuous exponential weights against strong adversaries, while \citet{cheng2026adversarial} studied adversarial bandit linear optimization over convex action sets with globally bounded perturbations against an oblivious adversary. Building on the principle of inducing a negative comparator-dependent FTRL penalty to cancel estimation or corruption errors \citep{foster2020adapting,zimmert2022return,liu2024corruption} and the monotone matrix envelope of \citet[Section~5.1]{liu2024corruption}, CRW-FTRL runs FTRL over the $K$-dimensional simplex with a radius-weighted entropy regularizer, while using a log-determinant barrier in an inner problem that selects the per-action radii for a normalized kernel estimator.

\paragraph{Multi-Armed Bandits with Adversarial Corruptions.}
\citet{lykouris2018stochastic} introduced stochastic multi-armed bandits with adversarial corruptions and proposed a corruption-agnostic algorithm whose regret incurs a multiplicative $\tilde O(KC)$ factor on top of the corruption-free regret. \citet{gupta2019better} improved this to a bound in which the corruption enters additively as $O(KC)$. In both works, the adversary corrupts the rewards of all arms before the learner draws its arm, which corresponds to the weak adversary in our terminology. For adversarial multi-armed bandits, \citet{yang2020adversarial} studied oblivious clean rewards with corruptions chosen after observing the pulled arm and obtained $O(\sqrt{KT\log K} + KC\log T)$ regret for a known corruption budget $C$, whereas \citet{zimmert2021tsallis} developed a best-of-both-worlds algorithm that is simultaneously optimal for stochastic and adversarial rewards and remains robust in corrupted stochastic environments. Corruption robustness has also been studied in episodic reinforcement learning \citep{lykouris2021corruption,wei2022model}. From the attacker's perspective, \citet{jun2018adversarial} and \citet{liu2019data} showed that a small corruption budget suffices to steer standard multi-armed bandit algorithms toward a target arm.

\paragraph{Stochastic Linear Bandits with Adversarial Corruptions.}
In the linear setting, \citet{garcelon2020adversarial} showed that standard linear contextual bandit algorithms are similarly vulnerable to targeted data poisoning attacks. On the defense side, \citet{li2019stochastic} obtained gap-dependent bounds for stochastic linear bandits against a weak adversary, measuring the corruption by its per-round maximum over actions. \citet{bogunovic2021stochastic} considered a strong adversary that observes the pulled arm and the noise before corrupting, proposed a robust phased elimination algorithm with $\tilde O(\sqrt{dT\log K} + d^{3/2}C)$ regret for known $C$, and proved that any algorithm must incur $\Omega(dC)$ regret even on an instance with $K = d$ orthogonal arms; see also \citet{bogunovic2020corruption} for Gaussian process bandits. Best-of-both-worlds algorithms for linear and linear contextual bandits that remain robust to corrupted stochastic rewards were developed by \citet{lee2021achieving} and \citet{kuroki2024best}. For linear contextual bandits, \citet{zhao2021linear} (under weak corruption) and \citet{ding2022robust} (under strong corruption) obtained corruption terms with a polynomial dependence on $T$, and \citet{he2022nearly} achieved the nearly optimal $\tilde O(d\sqrt{T} + dC)$ regret against a strong adversary by weighting each observation according to its uncertainty. This uncertainty-weighting idea has since been extended to nonlinear function classes \citep{ye2023corruption}, variance-aware bounds \citep{yu2025corruption}, and dueling bandits \citep{di2025nearly}. A closely related line of work studies misspecified linear bandits, where the reward deviates from a linear function by a bounded amount in every round \citep{lattimore2020learning,foster2020adapting,neu2020efficient}, and \citet{liu2024corruption} unified the weak ($C_\infty$) and strong ($C$) corruption measures across stochastic and adversarial linear bandits. Table~\ref{tab:comparison} compares our result with the most related work on linear bandits with corruptions.

\begin{table}[t]
\centering
\caption{Comparison of regret bounds for linear bandits with corruptions. ``Stoch.'' and ``Adv.'' indicate whether the reward vectors $\btheta_t$ are fixed or chosen by an adaptive adversary. $C = \sum_{t=1}^T |c_t|$ is the strong corruption measure and $C_\infty \ge C$ is the weak measure of \citet{liu2024corruption}. $K$ is the number of actions, $\tilde O(\cdot)$ and $\tilde\Omega(\cdot)$ hide logarithmic factors, and the lower bounds hold in the nontrivial regime where the stated bound is at most $O(T)$.}\label{tab:comparison}
\small
\setlength{\tabcolsep}{5pt}
\renewcommand{\arraystretch}{1.25}
\begin{tabular}{llcl}
\toprule
Algorithm & Rewards & Measure & Regret \\
\midrule
Robust Phased Elimination \citep{bogunovic2021stochastic} & Stoch. & $C$ & $\tilde O(\sqrt{dT\log K} + d^{3/2}C)$ \\
CW-OFUL \citep{he2022nearly} & Stoch. & $C$ & $\tilde O(d\sqrt{T} + dC)$ \\
Randomized Phased Elimination \citep{liu2024corruption} & Stoch. & $C_\infty$ & $\tilde O(d\sqrt{T} + \sqrt{d}C_\infty)$ \\
FTRL with log-det \citep{liu2024corruption} & Adv. & $C_\infty$ & $\tilde O(d\sqrt{T} + \sqrt{d}C_\infty)$ \\
Continuous Exp.\ Weights \citep{liu2024corruption} & Adv. & $C$ & $\tilde O(\sqrt{d^3T} + dC)$ \\
\rowcolor{LightCyan}
CRW-FTRL (Algorithm~\ref{alg:main}, ours), $|\cA| = K$ & Adv. & $C$ & $\tilde O(\sqrt{dT\log K} + dC)$ \\
\rowcolor{LightCyan}
CRW-FTRL on a cover (ours) & Adv. & $C$ & $\tilde O(d\sqrt{T} + dC)$ \\
\midrule
Lower bound \citep{dani2007price,lattimore2020learning} & --- & $C_\infty$ & $\tilde\Omega(d\sqrt{T} + \sqrt{d}C_\infty)$ \\
Lower bound \citep{dani2007price,bogunovic2021stochastic} & --- & $C$ & $\Omega(d\sqrt{T} + dC)$ \\
\bottomrule
\end{tabular}
\end{table}

\section{Problem Setting}\label{sec:setting}

\paragraph{Adversarial Linear Bandits with Strong Corruption.}
Let $\cA = \{\ab_1, \dots, \ab_K\} \subset \RR^d$ be a finite action set known to the learner. The interaction between the learner and an adaptive adversary proceeds over $T$ rounds. In each round $t \in [T]$:
\begin{enumerate}[leftmargin=*]
  \item Based on the interaction history before round $t$, the adversary chooses a reward vector $\btheta_t \in \RR^d$, which defines the \emph{clean reward} $\ab_i^\top \btheta_t$ of each action $i \in [K]$, while the learner simultaneously chooses a sampling distribution $\pb_t \in \Delta_K$ and draws an action $I_t \sim \pb_t$.
  \item A zero-mean noise variable $\xi_t \in \RR$ is realized, yielding the clean observation $\ab_{I_t}^\top \btheta_t + \xi_t$.
  \item After observing both the realized action $I_t$ and the clean observation $\ab_{I_t}^\top \btheta_t + \xi_t$, the adversary chooses a corruption $c_t \in \RR$, and the learner observes only the corrupted reward
  \begin{align*}
    y_t = \ab_{I_t}^\top \btheta_t + \xi_t + c_t .
  \end{align*}
\end{enumerate}
We impose the following regularity conditions on the action set, noise, and corruption budget.

\begin{assumption}[{Actions and Clean Rewards}]\label{ass:actions}
The action set $\cA = \{\ab_1, \dots, \ab_K\} \subset \RR^d$ is finite and its affine hull is $\RR^d$. Almost surely, $|\ab_i^\top\btheta_t| \le 1$ for all $i \in [K]$ and $t \in [T]$.
\end{assumption}

\begin{assumption}[{Filtration, Noise, and Corruption Budget}]\label{ass:protocol}
Let $\cF_t$ be the $\sigma$-field generated by $\{\btheta_s, I_s, \xi_s, c_s\}_{s=1}^t$, together with $\btheta_{t+1}$ and the internal randomness of the environment and the adversary prior to sampling $I_{t+1}$. For all $t \in [T]$:
\begin{enumerate}
  \item[(a)] $\btheta_t$ and $\pb_t$ are $\cF_{t-1}$-measurable, and $\PP(I_t = j \mid \cF_{t-1}) = p_{tj}$ for all $j \in [K]$.
  \item[(b)] $|\xi_t| \le 1$ almost surely, and $\EE[\xi_t \mid \cF_{t-1}, I_t] = 0$.
  \item[(c)] $|c_t| \le 1$ and $\sum_{t=1}^T |c_t| \le C$ almost surely, where $C \ge 0$ is the total corruption budget.
  \item[(d)] The learner knows the action set $\cA$, the horizon $T$, and the corruption budget $C$.
\end{enumerate}
\end{assumption}

Under Assumptions~\ref{ass:actions} and~\ref{ass:protocol}, the clean observation and the corrupted observation are bounded almost surely by $\abr{\ab_{I_t}^\top\btheta_t + \xi_t} \le 2$ and $\abr{y_t} \le 3$. The learner's goal is to minimize the realized regret evaluated on the clean rewards.

\begin{definition}[{Regret}]\label{def:regret}
The realized regret after $T$ rounds is defined as
\begin{align*}
  \mathrm{Reg}_T \coloneqq \max_{i \in [K]} \sum_{t=1}^T (\ab_i - \ab_{I_t})^\top\btheta_t .
\end{align*}
\end{definition}

\begin{remark}[Discussion of the Setting]\label{rem:setting}
We make three remarks regarding the problem formulation:
\begin{itemize}[leftmargin=*]
  \item \emph{Strong vs.\ weak corruption:}
  We call the adversary \emph{strong} because $c_t$ is chosen after observing the drawn action $I_t$ and clean observation $\ab_{I_t}^\top \btheta_t + \xi_t$ \citep{bogunovic2021stochastic,he2022nearly,liu2024corruption}. Under \emph{weak} corruption \citep{lykouris2018stochastic,gupta2019better,liu2024corruption}, by contrast, the adversary must commit to a corruption function $c_t : [K] \to \RR$ before $I_t \sim \pb_t$ is drawn. A strong adversary can spend its budget $C$ selectively by remaining silent in most rounds and corrupting only rounds in which a rarely sampled action ($p_{t,I_t} \ll 1$) is drawn and its reward estimate carries a large weight.
  \item \emph{Non-reducibility to uncorrupted adversarial linear bandits:}
  Even though $\btheta_t$ is already chosen by an adaptive adversary, strong corruption $c_t$ cannot be absorbed into $\btheta_t$ for two reasons: (i) $c_t$ depends on the realized action $I_t$, so the corrupted observation $y_t$ breaks the linear reward structure $\ab_{I_t}^\top \btheta'_t + \xi_t$ for any $\cF_{t-1}$-measurable vector $\btheta'_t \in \RR^d$; and (ii) $\mathrm{Reg}_T$ is evaluated on the clean rewards $\ab_i^\top \btheta_t$ rather than the corrupted observations.
  \item \emph{Scope of the assumptions:}
  Setting $\btheta_t \equiv \btheta$ recovers stochastic linear bandits under strong corruption, and setting $C = 0$ recovers adversarial linear bandits with noisy feedback. The full affine span condition in Assumption~\ref{ass:actions} is without loss of generality: if $\cA$ has affine dimension $d' < d$, the lifted actions $\zb_i = (\ab_i^\top, 1)^\top \in \RR^{d+1}$ span a $(d'+1)$-dimensional linear subspace of $\RR^{d+1}$, and expressing $\zb_i$ in a basis of that subspace (preserving the constant coordinate $1$) reduces $d$ to $d'$ while keeping every clean reward $\ab_i^\top \btheta_t$ linear in the reduced lifted coordinates. The per-round bound $|c_t| \le 1$ is also for convenience: since $|\ab_{I_t}^\top \btheta_t + \xi_t| \le 2$, clipping $y_t$ to $[-2, 2]$ ensures an effective per-round corruption of magnitude at most $\min\{|c_t|, 4\}$, affecting the regret bound only through constants.
\end{itemize}
\end{remark}

\section{Algorithm}\label{sec:algorithm}
Our algorithm, CRW-FTRL (Algorithm~\ref{alg:main}), maintains a sampling distribution over the $K$ actions and updates it by follow-the-regularized-leader (FTRL) on estimates of the clean rewards. Before play, it fixes a reference distribution over the actions and a change of coordinates computed from it (Section~\ref{sec:design}). During play, each round has two parts (Section~\ref{sec:ftrl-algo}): the learner computes the sampling distribution from the past estimates, and after observing the reward of the sampled action it forms an estimate of the clean reward of every action.

\subsection{Reference Design and Whitened Actions}\label{sec:design}
Before play begins, the algorithm performs a static change of coordinates that depends only on the action set $\cA$. We first lift each action $\ab_i \in \RR^d$ to $\zb_i \coloneqq (\ab_i^\top, 1)^\top \in \RR^{d+1}$ by appending a constant coordinate $1$. Lifting preserves linearity of the clean rewards ($\ab_i^\top \btheta_t = \zb_i^\top (\btheta_t^\top, 0)^\top$) while ensuring that every lifted action has a constant last coordinate $z_{i,d+1} = 1$, which will directly supply the uniform norm lower bound $\norm{\xb_i} \ge 1$ (and hence $r_{ti} \ge 1$) needed by our normalized kernel estimator without requiring any lower bound on $\norm{\ab_i}$. Moreover, since $\cA$ has full affine span in $\RR^d$ (Assumption~\ref{ass:actions}), the lifted actions $\zb_1, \dots, \zb_K$ linearly span $\RR^{d+1}$.

To put the lifted actions into a canonical position, we use a $D$-optimal ($G$-optimal) experimental design over $\{\zb_1, \dots, \zb_K\}$.

\begin{definition}[{Reference Design}]\label{def:design}
Let $\Hb(\pb) \coloneqq \sum_{i=1}^K p_i \zb_i \zb_i^\top$ for $\pb \in \Delta_K$, and let $\brho^\star \in \Delta_K$ be a maximizer of $\log\det \Hb(\pb)$. Set
\begin{align*}
  \brho \coloneqq \tfrac12\bigl(\brho^\star + \tfrac1K \mathbf 1\bigr), \qquad \xb_i \coloneqq \Hb(\brho)^{-1/2} \zb_i .
\end{align*}
\end{definition}

Because $\zb_1, \dots, \zb_K$ span $\RR^{d+1}$, the matrix $\Hb(\pb)$ is positive definite whenever $\pb$ has full support, so the maximum of $\log\det \Hb(\pb)$ over the compact simplex $\Delta_K$ is finite and attained at some design $\brho^\star$ with $\Hb(\brho^\star) \succ 0$. By the Kiefer--Wolfowitz equivalence theorem \citep{kiefer1960equivalence}, maximizing $\log\det \Hb(\pb)$ ($D$-optimality) is equivalent to minimizing the worst-case leverage score $\max_{i \in [K]} \zb_i^\top \Hb(\pb)^{-1} \zb_i$ ($G$-optimality), and the optimal design $\brho^\star$ satisfies
\begin{align*}
  \zb_i^\top \Hb(\brho^\star)^{-1} \zb_i \le d + 1 \qquad \text{for all } i \in [K] .
\end{align*}
The optimal design $\brho^\star$ cannot be used directly as a floor on the sampling probabilities because it may assign zero mass to some actions. The reference design $\brho$ in Definition~\ref{def:design} remedies this by mixing $\brho^\star$ with the uniform distribution $\frac1K \mathbf 1$. This mixture retains the benefits of both components:
\begin{itemize}[leftmargin=*]
  \item \emph{Probability floor:} Since half of the mass of $\brho$ is uniform, every action has reference probability at least $\rho_i \ge 1/(2K)$.
  \item \emph{Isotropy and bounded norms:} Under the whitened coordinates $\xb_i = \Hb(\brho)^{-1/2} \zb_i$, the actions are isotropic under $\brho$ and have bounded norms: since $\Hb(\brho) \succeq \frac12 \Hb(\brho^\star)$ implies $\Hb(\brho)^{-1} \preceq 2 \Hb(\brho^\star)^{-1}$,
  \begin{align*}
    \sum_{i=1}^K \rho_i \xb_i \xb_i^\top = \Ib, \qquad
    1 \le \norm{\xb_i}^2 = \zb_i^\top \Hb(\brho)^{-1} \zb_i \le 2 \zb_i^\top \Hb(\brho^\star)^{-1} \zb_i \le 2(d+1) \quad \text{for all } i \in [K] ,
  \end{align*}
  where the lower bound $\norm{\xb_i} \ge 1$ follows from the appended coordinate $1$.
  \item \emph{Linearity of rewards:} Defining the transformed parameter $\bvartheta_t \coloneqq \Hb(\brho)^{1/2}(\btheta_t^\top, 0)^\top$, the clean reward of every action $i$ is
  \begin{align*}
    \ab_i^\top \btheta_t = \zb_i^\top (\btheta_t^\top, 0)^\top = \xb_i^\top \bvartheta_t .
  \end{align*}
\end{itemize}
From now on, the algorithm uses $\cA$ only through the whitened actions $\xb_1, \dots, \xb_K$, and treats the reference design $\brho$ as known.

\subsection{Motivation: Controlling Sensitivity via Action Radii}\label{sec:motivation}
In each round $t$, our algorithm plays a sampling distribution $\pb_t \in \Delta_K$ computed by follow-the-regularized-leader (FTRL) on the cumulative reward estimates $\Sbb_{t-1} = \sum_{s < t} \ghb_s$, draws an action $I_t \sim \pb_t$, and uses the observed reward $y_t$ to construct a new reward estimate vector $\ghb_t \in \RR^K$.

In standard EXP3 for linear bandits, the learner estimates the reward of action $i$ by
\begin{align*}
  \gh^{\mathrm{IW}}_t(i) = \xb_i^\top (\Gb^{\mathrm{IW}}_t)^{-1} \xb_{I_t} y_t, \qquad \text{where } \Gb^{\mathrm{IW}}_t = \sum_{j=1}^K p_{tj} \xb_j \xb_j^\top .
\end{align*}
In round $t$, the corruption $c_t$ in $y_t = \ab_{I_t}^\top \btheta_t + \xi_t + c_t$ shifts this estimate by
\begin{align*}
  \Delta_t^{\mathrm{IW}}(i) = c_t\, \xb_i^\top (\Gb^{\mathrm{IW}}_t)^{-1} \xb_{I_t} .
\end{align*}
When an action $j$ has small sampling probability $p_{tj}$, its leverage score can be as large as
\begin{align*}
  \xb_j^\top (\Gb^{\mathrm{IW}}_t)^{-1} \xb_j \le \frac{1}{p_{tj}} ,
\end{align*}
with equality whenever $\xb_j$ is orthogonal to the remaining actions. Under weak corruption ($c_t$ fixed before drawing $I_t \sim \pb_t$), the factor $1/p_{tj}$ cancels against $\PP(I_t = j \mid \cF_{t-1}) = p_{tj}$ in expectation. Under strong corruption, however, the adversary sees $I_t$ before choosing $c_t$: it can set $c_t = 0$ unless an action $I_t = j$ with tiny probability $p_{tj}$ is drawn, in which case $|c_t| = 1$ perturbs $\gh^{\mathrm{IW}}_t(j)$ by up to $1/p_{tj}$.

The issue is that in $\gh^{\mathrm{IW}}_t(i) = \xb_i^\top (\Gb^{\mathrm{IW}}_t)^{-1} \xb_{I_t} y_t$, the weight on $y_t$ is bounded by Cauchy--Schwarz as
\begin{align*}
  \big|\xb_i^\top (\Gb^{\mathrm{IW}}_t)^{-1} \xb_j\big| \le \sqrt{\xb_i^\top (\Gb^{\mathrm{IW}}_t)^{-1} \xb_i}\,\sqrt{\xb_j^\top (\Gb^{\mathrm{IW}}_t)^{-1} \xb_j} ,
\end{align*}
where the second factor depends on the drawn action $j = I_t$ and can be as large as $1/\sqrt{p_{tj}}$. To eliminate this dependence on $I_t$, our algorithm assigns each action $i$ a \emph{radius} $r_{ti} = r_i(\Qb_t) \coloneqq \sqrt{\xb_i^\top \Qb_t \xb_i} \ge 1$ in round $t$ via a matrix $\Qb_t \succ 0$, and replaces $\xb_i^\top (\Gb^{\mathrm{IW}}_t)^{-1} \xb_{I_t} y_t$ with the \emph{reweighted kernel estimate} $k_t(i, I_t)\, y_t$, where
\begin{align*}
  k_t(i,j) \coloneqq \frac{\xb_i^\top \Gb_t^{-1} \xb_j}{r_{tj}}, \qquad
  \Gb_t \coloneqq \sum_{l=1}^K \frac{p_{tl}}{r_{tl}} \xb_l \xb_l^\top .
\end{align*}
Here $\Gb_t$ is defined with weights $p_{tl}/r_{tl}$ so that $k_t(i, I_t)\, y_t$ remains an unbiased estimator of the clean reward $\ab_i^\top \btheta_t = \xb_i^\top \bvartheta_t$ when $c_t = 0$: taking expectation over $I_t \sim \pb_t$,
\begin{align*}
  \EE\big[k_t(i, I_t)\, \xb_{I_t}^\top \bvartheta_t \;\big|\; \cF_{t-1}\big]
  = \sum_{j=1}^K p_{tj}\, \frac{\xb_i^\top \Gb_t^{-1} \xb_j}{r_{tj}}\, \xb_j^\top \bvartheta_t
  = \xb_i^\top \Gb_t^{-1} \bigg(\sum_{j=1}^K \frac{p_{tj}}{r_{tj}} \xb_j \xb_j^\top\bigg) \bvartheta_t
  = \xb_i^\top \bvartheta_t = \ab_i^\top \btheta_t .
\end{align*}
At the same time, dividing by $r_{tj}$ in $k_t(i,j)$ cancels the factor from the drawn action $j = I_t$: whenever the radii satisfy the envelope condition $\xb_l^\top \Gb_t^{-1} \xb_l \le r_{tl}^2$ for all $l \in [K]$, Cauchy--Schwarz gives
\begin{align*}
  |k_t(i,j)| = \frac{|\xb_i^\top \Gb_t^{-1} \xb_j|}{r_{tj}}
  \le \sqrt{\xb_i^\top \Gb_t^{-1} \xb_i}\cdot \frac{\sqrt{\xb_j^\top \Gb_t^{-1} \xb_j}}{r_{tj}}
  \le r_{ti}\cdot \frac{r_{tj}}{r_{tj}} = r_{ti} \qquad \text{for all } i, j \in [K] .
\end{align*}
Thus, whichever action $I_t$ is drawn, a corruption $c_t$ shifts action $i$'s estimate $k_t(i, I_t)\, y_t$ by at most $|c_t| r_{ti}$. Although $r_{ti}$ itself can be large when $p_{ti}$ is small, the bound now depends only on $i$: if the radii are nondecreasing ($r_{1i} \le \dots \le r_{Ti}$), the total corruption accumulated on any comparator $\ub \in \Delta_K$ over $T$ rounds is bounded by
\begin{align*}
  \sum_{t=1}^T \sum_{i=1}^K u_i\, c_t k_t(i, I_t) \le \sum_{t=1}^T |c_t| \sum_{i=1}^K u_i r_{ti} \le C \inner{\ub}{\rb_T} ,
\end{align*}
which is linear in the comparator's final radius $\inner{\ub}{\rb_T}$ and can therefore be canceled directly by a negative linear term $-\alpha \inner{\pb}{\rb_t}$ in the FTRL regularizer. Building on this idea, our algorithm consists of four components, which we detail in Section~\ref{sec:ftrl-algo}:
\begin{enumerate}[leftmargin=*]
  \item \emph{Floored simplex $\Pg \coloneqq \{\pb \in \Delta_K : p_i \ge \rho_i/(2T)\}$:} Enforcing $p_i \ge \rho_i/(2T)$ ensures $\log(1/p_i) \le \log(4KT)$ for all $i \in [K]$.
  \item \emph{Radius-weighted regularizer:} Weighting coordinate $i$ of the negative entropy by $r_{ti}$ keeps FTRL stable for estimates of scale $r_{ti}$, while a linear term $-\alpha \sum_i r_{ti} p_i$ contributes $-\alpha \inner{\ub}{\rb_T}$ at any comparator $\ub$ to cancel its estimation and corruption errors.
  \item \emph{Monotone log-det inner problem:} Choosing $\Qb_t \succeq \Qb_{t-1}$ jointly with $\pb_t$ enforces $\xb_l^\top \Gb_t^{-1} \xb_l \le r_{tl}^2$ via first-order optimality while bounding the learner's regularizer drift by $O(\kappa d\log(dT))$.
  \item \emph{Normalized kernel estimator:} Centering $k_t(i,j)$ to $\kt_t(i,j)$ keeps $\kt_t$ unbiased with $|\kt_t(i,j)| \le 3r_{ti}$, and enforces $\sum_i p_{ti}\kt_t(i, I_t) = 1$ so that $c_t$ enters the learner's own reward with weight $1$.
\end{enumerate}

\subsection{CRW-FTRL: Corruption-Robust Radius-Weighted FTRL}\label{sec:ftrl-algo}

\paragraph{Floored Simplex.}
The learner optimizes $\pb_t$ over the floored simplex
\begin{align*}
  \Pg \coloneqq \big\{\pb \in \Delta_K : p_i \ge \rho_i/(2T) \text{ for all } i \in [K]\big\} .
\end{align*}
Since $\rho_i \ge 1/(2K)$ (Section~\ref{sec:design}), every $\pb \in \Pg$ satisfies $p_i \ge 1/(4KT)$ and hence $0 \le \log(1/p_i) \le \log(4KT)$. Meanwhile, mixing any comparator with $\frac{1}{2T}\brho$ changes its clean reward by at most $1/T$ per round, costing at most $1$ over $T$ rounds.

\paragraph{Regularizer Weighted by Radii.}
Given cumulative reward estimates $\Sbb_{t-1} = \sum_{s=1}^{t-1} \ghb_s$ and a convex regularizer $\psi_t: \Pg \to \RR$, the follow-the-regularized-leader (FTRL) rule selects $\pb_t \in \Pg$ by maximizing the cumulative estimated reward minus the regularizer:
\begin{align*}
  \pb_t = \operatorname*{arg\,max}_{\pb \in \Pg} \big\{\inner{\Sbb_{t-1}}{\pb} - \psi_t(\pb)\big\} .
\end{align*}
For any comparator $\ub \in \Pg$, standard FTRL analysis (Section~\ref{sec:tool-ftrl}) bounds the regret on the estimated rewards by three terms:
\begin{align*}
  \sum_{t=1}^T \inner{\ub - \pb_t}{\ghb_t}
  \;\lesssim\;
  \underbrace{\psi_T(\ub) - \psi_0(\pb_1)}_{\text{penalty at comparator } \ub}
  \;+\; \underbrace{\sum_{t=1}^T \big(\psi_{t-1}(\pb_t) - \psi_t(\pb_t)\big)}_{\text{regularizer drift at } \pb_t}
  \;+\; \underbrace{\sum_{t=1}^T \sum_{i=1}^K \frac{\gh_t(i)^2}{\partial_{p_i}^2 \psi_t(\pb_t)}}_{\text{stability term}} .
\end{align*}
In standard EXP3, the regularizer is the negative entropy $\psi(\pb) = \eta^{-1}\sum_{i=1}^K p_i \log p_i$ with learning rate $\eta > 0$, which has second derivative $\partial_{p_i}^2 \psi(\pb) = 1/(\eta p_i)$. In our setting, given the radius matrix $\Qb_t \succ 0$ with radii $r_{ti} = r_i(\Qb_t)$, we use $\psi_t = \psi_{\Qb_t}$, where for any $\Qb \succ 0$ and offset $\alpha > 0$ we define
\begin{align*}
  \psi_{\Qb}(\pb) \coloneqq \sum_{i=1}^K r_i(\Qb)\big(\eta^{-1} p_i\log p_i - \alpha p_i\big)
  = -\sum_{i=1}^K r_i(\Qb)\, h(p_i), \qquad
  h(s) \coloneqq s\,\big(\alpha + \eta^{-1}\log(1/s)\big) .
\end{align*}
Each part of $\psi_{\Qb_t}(\pb)$ is designed to control one term in the FTRL bound above:
\begin{enumerate}[leftmargin=*]
  \item \emph{Controlling the stability term via $r_{ti} p_i \log p_i$:} Multiplying coordinate $i$'s negative entropy by $r_{ti}$ gives second derivative $\partial_{p_i}^2 \psi_{\Qb_t}(\pb_t) = r_{ti}/(\eta p_{ti})$. Substituting this into the stability term yields
  \begin{align*}
    \sum_{i=1}^K \frac{\gh_t(i)^2}{\partial_{p_i}^2 \psi_{\Qb_t}(\pb_t)} = \eta \sum_{i=1}^K \frac{p_{ti}}{r_{ti}}\, \gh_t(i)^2 ,
  \end{align*}
  which divides $\gh_t(i)^2$ by $r_{ti}$ to compensate for the $O(r_{ti})$ magnitude of $\gh_t(i)$.
  \item \emph{Making the comparator penalty negative via $-\alpha r_{ti} p_i$:} Since $u_i \log u_i \le 0$ for every $\ub \in \Pg$, the comparator penalty $\psi_T(\ub) = \psi_{\Qb_T}(\ub)$ is negative:
  \begin{align*}
    \psi_{\Qb_T}(\ub) = \sum_{i=1}^K r_{Ti}\big(\eta^{-1} u_i \log u_i - \alpha u_i\big) \le -\alpha \sum_{i=1}^K u_i r_{Ti} = -\alpha \inner{\ub}{\rb_T} .
  \end{align*}
  Notice that FTRL bounds the regret on the \emph{estimated} rewards $\ghb_t$, whereas true regret is measured on the clean rewards $\ab_i^\top \btheta_t$. Passing from $\inner{\ub}{\ghb_t}$ to $\ub$'s true clean reward incurs two estimation errors that both scale with the estimator's range $O(r_{Ti})$: a \emph{corruption error} $O(C\inner{\ub}{\rb_T})$ from the adversarial corruptions $c_t$ in $y_t$, and a \emph{concentration error} $O\big(\eta^{-1}\log(KT/\delta)\,\inner{\ub}{\rb_T}\big)$ from the random fluctuations of the clean estimates around their mean $\ab_i^\top \btheta_t$ (due to sampling $I_t \sim \pb_t$ and noise $\xi_t$). Taking $\alpha \ge 3C + O\big(\eta^{-1}\log(KT/\delta)\big)$ allows $-\alpha\inner{\ub}{\rb_T}$ to cancel both errors directly. This follows the general principle of inducing a negative terminal comparator penalty (or fixed-point bias) in FTRL to offset estimation or corruption errors \citep{foster2020adapting,zimmert2022return,liu2024corruption}, adapted here to a radius-weighted entropy on the simplex so that strong corruption is controlled through the radii $r_{ti}$.
  \item \emph{Bounding the regularizer drift via $h(p_i)$:} Since $0 \le \log(1/p_i) \le \log(4KT)$ for all $\pb \in \Pg$, setting $\kappa \coloneqq \alpha + \eta^{-1}\log(4KT)$ sandwiches $h(p_i)$ within a constant factor of $p_i$:
  \begin{align*}
    \alpha p_i \le h(p_i) \le \kappa p_i \qquad \text{for all } \pb \in \Pg \text{ and } i \in [K] .
  \end{align*}
  Consequently, when $r_{ti} \ge r_{t-1,i}$, the regularizer drift in round $t$ is bounded by
  \begin{align*}
    \psi_{\Qb_{t-1}}(\pb_t) - \psi_{\Qb_t}(\pb_t) = \sum_{i=1}^K h(p_{ti})(r_{ti} - r_{t-1,i}) \le \kappa \sum_{i=1}^K p_{ti}(r_{ti} - r_{t-1,i}) ,
  \end{align*}
  where the small probability $p_{ti}$ multiplies the radius increment $r_{ti} - r_{t-1,i}$.
\end{enumerate}

\paragraph{Choosing the Radii via a Monotone Log-Det Problem.}
In round $t$, the radii $r_{ti}$ must satisfy the envelope condition $\xb_i^\top \Gb_t^{-1} \xb_i \le r_{ti}^2$ and monotonicity $r_{ti} \ge r_{t-1,i}$ for all $i \in [K]$, while keeping the regularizer drift $\psi_{\Qb_{t-1}}(\pb_t) - \psi_{\Qb_t}(\pb_t)$ as small as possible. This motivates the following three-step design (see also \citet[Section~5.1]{liu2024corruption} for a related monotone matrix envelope and log-determinant drift bound under weak corruption):
\begin{enumerate}[leftmargin=*]
  \item \emph{Matrix parametrization of the radii:} Observing that the left-hand side of the envelope condition $\xb_i^\top \Gb_t^{-1} \xb_i \le r_{ti}^2$ is a quadratic form of the $(d+1)\times(d+1)$ matrix $\Gb_t^{-1}$, we parametrize the upper bound $r_{ti}^2$ as a quadratic form of a $(d+1)\times(d+1)$ matrix $\Qb_t \succ 0$:
  \begin{align*}
    r_{ti} = r_i(\Qb_t) \coloneqq \sqrt{\xb_i^\top \Qb_t \xb_i} .
  \end{align*}
  This reduces the $K$ arm-wise monotonicity and envelope conditions to two matrix inequalities in $\RR^{(d+1)\times(d+1)}$:
  \begin{align*}
    \Qb_t \succeq \Qb_{t-1} \;\; (\implies r_{ti} \ge r_{t-1,i}),
    \qquad
    \Gb_t = \sum_{i=1}^K \frac{p_{ti}}{r_i(\Qb_t)} \xb_i \xb_i^\top \succeq \Qb_t^{-1} \;\; (\implies \xb_i^\top \Gb_t^{-1} \xb_i \le r_{ti}^2) .
  \end{align*}

  \item \emph{Log-determinant barrier for the envelope condition:}
  For a given $\pb_t \in \Pg$, choosing $\Qb_t \succeq \Qb_{t-1}$ faces a trade-off:
  making $\Qb_t$ larger hurts the learner through the regularizer drift
  \begin{align*}
    \psi_{\Qb_{t-1}}(\pb_t) - \psi_{\Qb_t}(\pb_t) = \sum_{i=1}^K h(p_{ti})\big(r_i(\Qb_t) - r_i(\Qb_{t-1})\big) ,
  \end{align*}
  so we want to minimize $\sum_{i=1}^K h(p_{ti}) r_i(\Qb) = -\psi_{\Qb}(\pb_t)$; at the same time, $\Qb_t$ cannot be too small because it must satisfy $\Gb_t \succeq \Qb_t^{-1}$. To enforce this lower bound on $\Qb_t$ while minimizing $-\psi_{\Qb}(\pb_t)$, we subtract the standard log-determinant barrier $\frac{\kappa}{2}\log\det \Qb$ and define
  \begin{align*}
    J(\Qb, \pb) \coloneqq \sum_{i=1}^K h(p_i)\, r_i(\Qb) - \frac{\kappa}{2}\log\det \Qb
    = -\psi_{\Qb}(\pb) - \frac{\kappa}{2}\log\det \Qb .
  \end{align*}
  Since $\nabla_{\Qb} r_i(\Qb) = \frac{1}{2r_i(\Qb)} \xb_i \xb_i^\top$ and $\nabla_{\Qb} \log\det \Qb = \Qb^{-1}$, the gradient of $J(\Qb, \pb)$ with respect to $\Qb$ is
  \begin{align*}
    \nabla_{\Qb} J(\Qb, \pb) = \frac{1}{2}\sum_{i=1}^K \frac{h(p_i)}{r_i(\Qb)} \xb_i \xb_i^\top - \frac{\kappa}{2} \Qb^{-1} .
  \end{align*}
  Without the monotonicity constraint $\Qb \succeq \Qb_{t-1}$, setting $\nabla_{\Qb} J(\Qb, \pb) = 0$ yields $\Qb^{-1} = \sum_{i=1}^K \frac{h(p_i)/\kappa}{\sqrt{\xb_i^\top \Qb \xb_i}} \xb_i \xb_i^\top$, which is the classical determinant optimization characterization of $\ell_1$ Lewis weights \citep{lewis1978finite,cohen2015lp} for the rescaled rows $\frac{h(p_i)}{\kappa}\xb_i^\top$. Minimizing $J(\Qb, \pb_t)$ over the constrained cone $\Qb \succeq \Qb_{t-1}$ extends this geometry to maintain nondecreasing radii and achieves both goals at once (Lemma~\ref{lem:envelope}): first-order optimality $2\nabla_{\Qb} J(\Qb_t, \pb_t) \succeq 0$ (together with $h(p_{ti}) \le \kappa p_{ti}$) enforces $\Gb_t \succeq \Qb_t^{-1}$, while $J(\Qb_t, \pb_t) \le J(\Qb_{t-1}, \pb_t)$ bounds the per-round drift $\psi_{\Qb_{t-1}}(\pb_t) - \psi_{\Qb_t}(\pb_t)$ by $\frac{\kappa}{2}(\log\det \Qb_t - \log\det \Qb_{t-1})$.

  \item \emph{Joint saddle-point update for $(\pb_t, \Qb_t)$:}
  Finally, $\pb_t$ and $\Qb_t$ are coupled: FTRL needs $\psi_{\Qb_t}$ to select $\pb_t$, whereas the barrier problem $\min_{\Qb \succeq \Qb_{t-1}} J(\Qb, \pb_t)$ needs $\pb_t$ to select $\Qb_t$. Because the barrier term $-\frac{\kappa}{2}\log\det \Qb$ in $J(\Qb, \pb) = -\psi_{\Qb}(\pb) - \frac{\kappa}{2}\log\det \Qb$ does not depend on $\pb$, maximizing $\inner{\Sbb_{t-1}}{\pb} - \psi_{\Qb}(\pb)$ over $\pb$ is equivalent to maximizing $\inner{\Sbb_{t-1}}{\pb} + J(\Qb, \pb)$. Starting from $\Qb_0 = \Ib$, the learner therefore computes $(\pb_t, \Qb_t)$ jointly via the max-min problem
  \begin{align*}
    \pb_t = \operatorname*{arg\,max}_{\pb \in \Pg} \Big\{\inner{\Sbb_{t-1}}{\pb} + \min_{\Qb \succeq \Qb_{t-1}} J(\Qb,\pb)\Big\}, \qquad
    \Qb_t = \operatorname*{arg\,min}_{\Qb \succeq \Qb_{t-1}} J(\Qb, \pb_t) ,
  \end{align*}
  where the inner objective $J(\Qb, \pb)$ is strictly geodesically convex in $\Qb$ along the matrix geometric mean path and the outer objective is strongly concave in $\pb$ (Lemmas~\ref{lem:inner} and~\ref{lem:outer}(a)). By a first-order optimality argument (Lemmas~\ref{lem:frozen} and~\ref{lem:outer}(b)), at the minimizer $\Qb_t$, the outer solution $\pb_t$ is simultaneously the exact FTRL iterate $\arg\max_{\pb \in \Pg}\{\inner{\Sbb_{t-1}}{\pb} - \psi_{\Qb_t}(\pb)\}$ for the frozen regularizer $\psi_{\Qb_t}$.
\end{enumerate}

\paragraph{Normalized Kernel Reward Estimates.}
Having computed $(\pb_t, \Qb_t)$ and the radii $r_{ti} = r_i(\Qb_t)$, the learner draws $I_t \sim \pb_t$, observes $y_t$, and constructs the reward estimate vector $\ghb_t \in \RR^K$ by
\begin{align*}
  \gh_t(i) \coloneqq \kt_t(i, I_t)\, y_t + 4\eta\, v_{ti},
\end{align*}
where, with the radius-weighted covariance matrix $\Gb_t \coloneqq \sum_{l=1}^K (p_{tl}/r_{tl})\, \xb_l \xb_l^\top$ and the reweighted kernel $k_t(i,j) \coloneqq \xb_i^\top \Gb_t^{-1} \xb_j / r_{tj}$ from Section~\ref{sec:motivation}, the \emph{normalized kernel} $\kt_t(i,j)$ and \emph{variance bonus} $v_{ti}$ are defined by
\begin{align}\label{eq:kt-vt}
  \kt_t(i,j) \coloneqq k_t(i,j) - \frac{r_{ti}}{\sum_{m=1}^K p_{tm} r_{tm}}\bigg(\sum_{l=1}^K p_{tl} k_t(l,j) - 1\bigg), \qquad
  v_{ti} \coloneqq \frac{1}{r_{ti}}\sum_{j=1}^K p_{tj} \kt_t(i,j)^2 .
\end{align}
Let us see how $\kt_t(i,j)$ and $v_{ti}$ work (Lemma~\ref{lem:kernel}):
\begin{enumerate}[leftmargin=*]
  \item \emph{Unbiasedness of the normalized kernel $\kt_t$:}
  When $c_t = 0$, we saw in Section~\ref{sec:motivation} that $k_t(l, I_t)\, y_t$ is an unbiased estimator of action $l$'s clean reward $\ab_l^\top \btheta_t$. Taking the $\pb_t$-weighted average over $l \in [K]$ shows that $\big(\sum_{l=1}^K p_{tl} k_t(l, I_t)\big) y_t$ is an unbiased estimator of the learner's expected reward $\sum_{l=1}^K p_{tl} \ab_l^\top \btheta_t$. At the same time, since $I_t \sim \pb_t$, the raw observation $1 \cdot y_t$ has the exact same conditional mean $\sum_{j=1}^K p_{tj} \ab_j^\top \btheta_t$. Therefore, their difference has zero conditional mean when $c_t = 0$:
  \begin{align*}
    \EE\bigg[\bigg(\sum_{l=1}^K p_{tl} k_t(l, I_t) - 1\bigg) y_t \;\bigg|\; \cF_{t-1}\bigg]
    = \sum_{l=1}^K p_{tl}\, \EE\big[k_t(l, I_t)\, y_t \mid \cF_{t-1}\big] - \EE\big[y_t \mid \cF_{t-1}\big] = 0 .
  \end{align*}
  Since the prefactor $r_{ti} / (\sum_m p_{tm} r_{tm})$ is $\cF_{t-1}$-measurable, subtracting this zero-mean term keeps $\kt_t(i, I_t)\, y_t$ unbiased for $\ab_i^\top \btheta_t$:
  \begin{align*}
    \EE\big[\kt_t(i, I_t)\, y_t \mid \cF_{t-1}\big] = \EE\big[k_t(i, I_t)\, y_t \mid \cF_{t-1}\big] = \ab_i^\top \btheta_t .
  \end{align*}

  \item \emph{Unit column sum and $O(r_{ti})$ bound:}
  Although $\big(\sum_{l=1}^K p_{tl} k_t(l, I_t)\big) y_t$ and $y_t$ have the same expectation when $c_t = 0$, they are not equal point-wise because dividing by $r_{tj}$ in $k_t(l,j)$ makes $\sum_{l=1}^K p_{tl} k_t(l,j) \neq 1$ in general. Subtracting the correction term in $\kt_t(i,j)$ distributes the discrepancy $\sum_{l=1}^K p_{tl} k_t(l,j) - 1$ across coordinates $i$ in proportion to $r_{ti}$, so that the $\pb_t$-weighted column sum of $\kt_t$ is identically $1$ for every $j \in [K]$:
  \begin{align*}
    \sum_{i=1}^K p_{ti} \kt_t(i,j)
    = \sum_{i=1}^K p_{ti} k_t(i,j) - \frac{\sum_{i=1}^K p_{ti} r_{ti}}{\sum_{m=1}^K p_{tm} r_{tm}}\bigg(\sum_{l=1}^K p_{tl} k_t(l,j) - 1\bigg) = 1 .
  \end{align*}
  Consequently, $\sum_{i=1}^K p_{ti} \kt_t(i, I_t)\, y_t = y_t$ holds point-wise for every drawn action $I_t$, ensuring that any corruption $c_t$ in $y_t$ enters the learner's own weighted estimate $\inner{\pb_t}{\ghb_t}$ with weight $1$. Moreover, since $|k_t(l,j)| \le r_{tl}$ and $r_{tm} \ge 1$ (so $\sum_m p_{tm} r_{tm} \ge 1$), the correction changes $k_t(i,j)$ by at most $2r_{ti}$:
  \begin{align*}
    |\kt_t(i,j)| \le |k_t(i,j)| + r_{ti}\,\frac{\sum_{l=1}^K p_{tl} |k_t(l,j)| + 1}{\sum_{m=1}^K p_{tm} r_{tm}}
    \le r_{ti} + r_{ti}\,\frac{\sum_{l=1}^K p_{tl} r_{tl} + 1}{\sum_{m=1}^K p_{tm} r_{tm}} \le 3 r_{ti} .
  \end{align*}

  \item \emph{Variance bonus $4\eta v_{ti}$:}
  As in high-probability adversarial bandit algorithms such as EXP3.P \citep{auer2002nonstochastic,beygelzimer2011contextual,lattimore2020bandit}, we augment the unbiased estimate with an optimistic variance bonus $+4\eta v_{ti}$. By a Freedman-type exponential-moment bound (Section~\ref{sec:tool-conc}), this bonus cancels the per-action cumulative variance $4\eta \sum_{t=1}^T v_{ti}$ (which can be large for rarely sampled actions), while shifting the learner's own weighted estimate $\inner{\pb_t}{\ghb_t}$ by only $4\eta \inner{\pb_t}{\vb_t} \le 8\eta$ per round (Lemma~\ref{lem:kernel}(e)).
\end{enumerate}

\begin{algorithm}
\caption{CRW-FTRL: Corruption-Robust Radius-Weighted FTRL}\label{alg:main}
\begin{algorithmic}[1]
\REQUIRE learning rate $\eta \in (0, 1/12]$, coefficients $\alpha > 0$ and $\kappa = \alpha + \eta^{-1}\log(4KT)$; the whitened actions $\xb_i$ and the design $\brho$ of Definition~\ref{def:design}
\STATE initialize $\Qb_0 = \Ib$ and $\Sbb_0 = \mathbf 0$
\FOR{$t = 1, \dots, T$}
  \STATE $\pb_t = \operatorname*{arg\,max}_{\pb \in \Pg} \big\{\inner{\Sbb_{t-1}}{\pb} + \min_{\Qb \succeq \Qb_{t-1}} J(\Qb,\pb)\big\}$ \hfill \texttt{// FTRL step}
  \STATE $\Qb_t = \operatorname*{arg\,min}_{\Qb \succeq \Qb_{t-1}} J(\Qb, \pb_t)$ and $r_{ti} = r_i(\Qb_t)$ \hfill \texttt{// radius update}
  \STATE $\Gb_t = \sum_{l=1}^K (p_{tl}/r_{tl})\, \xb_l \xb_l^\top$, $k_t(i,j) = \xb_i^\top \Gb_t^{-1} \xb_j / r_{tj}$, and compute $(\kt_t, \vb_t)$ via~\eqref{eq:kt-vt} \hfill \texttt{// kernel}
  \STATE draw $I_t \sim \pb_t$ and observe $y_t$
  \STATE $\gh_t(i) = \kt_t(i, I_t)\, y_t + 4\eta\, v_{ti}$ for all $i$, and $\Sbb_t = \Sbb_{t-1} + \ghb_t$ \hfill \texttt{// reward estimates}
\ENDFOR
\end{algorithmic}
\end{algorithm}

The complete procedure of CRW-FTRL is summarized in Algorithm~\ref{alg:main}. The optimizers in lines~3 and~4 exist and are unique (Lemmas~\ref{lem:inner} and~\ref{lem:outer}), and the matrix $\Gb_t = \sum_{l=1}^K (p_{tl}/r_{tl})\, \xb_l \xb_l^\top$ in line~5 is positive definite because $p_{tl} > 0$ and $r_{tl} > 0$ for every $l$ and the vectors $\xb_1, \dots, \xb_K$ span $\RR^{d+1}$.

\section{Main Results}\label{sec:main-results}
Our main result bounds the realized regret of CRW-FTRL (Algorithm~\ref{alg:main}) with high probability.

\begin{theorem}\label{thm:main}
Suppose Assumptions~\ref{ass:actions} and~\ref{ass:protocol} hold. Fix $\eta \in (0, 1/12]$ and $\delta \in (0,1)$, let $L = \log(8KT/\delta)$, and run Algorithm~\ref{alg:main} with $\alpha = 3C + L/\eta$ and $\kappa = \alpha + \log(4KT)/\eta$. Then with probability at least $1 - \delta$,
\begin{align*}
  \mathrm{Reg}_T \le 3A_T C + \frac{2A_T L}{\eta} + 48\eta T + C + \sqrt{2T\log(2/\delta)} + 1 ,
\end{align*}
where $A_T = \sqrt{2(d+1)} + (d+1) \log\big(1 + 4\sqrt2\, (d+1)^{3/2} T^2\big) = O(d\log(dT))$.
\end{theorem}
\noindent\emph{Proof.} We give a proof overview in Section~\ref{sec:overview} and the complete proof in Appendix~\ref{app:app-main}.

Choosing the learning rate $\eta$ to balance $2A_T L/\eta$ against $48\eta T$ yields our main regret bound.

\begin{corollary}\label{cor:rate}
Under the conditions of Theorem~\ref{thm:main}, let $\eta = \min\{1/12,\ \sqrt{A_T L/(24T)}\}$. With probability at least $1-\delta$,
\begin{align*}
  \mathrm{Reg}_T = O\Big(dC\log(dT) + \sqrt{dT\log(dT)\log(KT/\delta)}\Big) = \tilde O\big(\sqrt{dT\log K} + dC\big).
\end{align*}
\end{corollary}
\noindent\emph{Proof.} See Appendix~\ref{app:app-main}.

\begin{remark}[Discussion of the Main Bound]\label{rem:main-discussion}
We highlight four key features of Theorem~\ref{thm:main} and Corollary~\ref{cor:rate}:
\begin{itemize}[leftmargin=*]
  \item \emph{Optimal $\tilde O(dC)$ strong-corruption cost and $\log K$ scaling:}
  The regret bound consists of an uncorrupted term $\tilde O(\sqrt{dT\log K})$, matching the best known rate for finite-arm linear bandits \citep{bubeck2012towards,zimmert2022return} with only logarithmic dependence on $K$, plus an additive corruption term $O(dC\log(dT)) = \tilde O(dC)$. Unlike weak corruption (where committing to $c_t$ before $I_t \sim \pb_t$ is drawn allows randomized exploration to average out the attack to $\tilde O(\sqrt{d}C_\infty)$ in terms of the weak measure $C_\infty$ of Section~\ref{sec:related} \citep{liu2024corruption}), strong corruption forces an $\Omega(dC)$ lower bound even in stochastic linear bandits \citep{bogunovic2021stochastic}, which our $\tilde O(dC)$ bound matches up to logarithmic factors.
  \item \emph{Matching stochastic linear bandits under strong corruption:}
  Crucially, the $\tilde O(\sqrt{dT\log K} + dC)$ bound holds when the clean rewards $\btheta_t$ are chosen by an adaptive adversary and $c_t$ is chosen by a \emph{strong} adversary after observing $I_t$ and the clean reward. Together with existing results for the uncorrupted and weak-corruption settings, our result shows that for fixed action sets, under both weak and strong corruption, adversarial linear bandits achieve the same minimax regret rates as stochastic linear bandits \citep{bogunovic2021stochastic,he2022nearly,liu2024corruption}.
  \item \emph{High-probability realized regret:}
  Theorem~\ref{thm:main} bounds the realized regret $\max_{i \in [K]} \sum_{t=1}^T (\ab_i - \ab_{I_t})^\top \btheta_t$ with high probability $1-\delta$ against an adaptive adversary, where the best action in hindsight is itself random. Setting $\delta = 1/T$ and using $\mathrm{Reg}_T \le 2T$ on the failure event immediately implies the same $\tilde O(\sqrt{dT\log K} + dC)$ bound for the expected regret $\EE[\mathrm{Reg}_T]$.
  \item \emph{Extension to infinite action sets via covering:}
  Because the number of actions enters Corollary~\ref{cor:rate} only through $\log K$, our result extends to any compact action set $\cA \subset \{\ab \in \RR^d : \norm{\ab} \le R\}$ when the reward vectors satisfy $\norm{\btheta_t} \le B$ almost surely (with $RB \le 1$): running Algorithm~\ref{alg:main} on a Euclidean $1/(BT)$-covering net $\cA_{\mathrm{net}} \subset \cA$ of size $K \le (1 + 2RBT)^d$ (augmented by at most $d+1$ affinely independent points of $\cA$ so that $\cA_{\mathrm{net}}$ shares the affine hull of $\cA$) incurs discretization error at most $\max_{\ab \in \cA}\min_{\ab' \in \cA_{\mathrm{net}}} \sum_{t=1}^T (\ab - \ab')^\top \btheta_t \le T \cdot \frac{1}{BT} \cdot B = 1$, satisfies $\log K = O(d\log T)$, and yields the minimax optimal $\tilde O(d\sqrt{T} + dC)$ regret bound.
\end{itemize}
\end{remark}

\section{Proof Overview}\label{sec:overview}
In this section, we present an overview of the proof of Theorem~\ref{thm:main} by decomposing the regret into four terms and bounding each term in turn. Throughout this section and the appendices, Algorithm~\ref{alg:main} is run with the parameters $\eta \in (0, 1/12]$, $L = \log(8KT/\delta)$, $\alpha = 3C + L/\eta$, and $\kappa = \alpha + \log(4KT)/\eta$ specified in Theorem~\ref{thm:main}. Complete proofs of Theorem~\ref{thm:main}, Corollary~\ref{cor:rate}, and all supporting technical lemmas are given in the appendices.

\subsection{Regret Decomposition}\label{sec:barrier}
Because the adversary is adaptive, the best action in hindsight $\argmax_{i \in [K]} \sum_{t=1}^T \ab_i^\top \btheta_t$ is random. We therefore fix an arbitrary comparator action $i^\star \in [K]$ and establish our high-probability bound simultaneously for all $i^\star \in [K]$ on a single event of probability at least $1-\delta$ that does not depend on $i^\star$.

Since the point mass $\eb_{i^\star}$ does not belong to the floored simplex $\Pg = \{\pb \in \Delta_K : p_i \ge \rho_i/(2T)\}$, we compare the learner against the smoothed comparator
\begin{align*}
  \ub \coloneqq \Big(1 - \frac{1}{2T}\Big) \eb_{i^\star} + \frac{1}{2T}\,\brho \in \Pg .
\end{align*}
Adding and subtracting $\sum_{i=1}^K u_i \ab_i^\top \btheta_t$, $\inner{\ub}{\ghb_t}$, and $\inner{\pb_t}{\ghb_t}$ decomposes the cumulative regret against $i^\star$ into four terms:
\begin{align}
  \sum_{t=1}^T (\ab_{i^\star} - \ab_{I_t})^\top\btheta_t
  ={}& \underbrace{\sum_{t=1}^T \Big(\ab_{i^\star}^\top\btheta_t - \sum_{i=1}^K u_i \ab_i^\top\btheta_t\Big)}_{T_0:\ \text{smoothing error}}
  + \underbrace{\sum_{t=1}^T \inner{\ghb_t}{\ub - \pb_t}}_{T_1:\ \text{FTRL regret}} \notag\\
  &+ \underbrace{\sum_{t=1}^T \Big(\sum_{i=1}^K u_i \ab_i^\top\btheta_t - \inner{\ub}{\ghb_t}\Big)}_{T_2:\ \text{comparator error}}
  + \underbrace{\sum_{t=1}^T \big(\inner{\pb_t}{\ghb_t} - \ab_{I_t}^\top\btheta_t\big)}_{T_3:\ \text{learner error}} . \label{eq:decomp}
\end{align}
Furthermore, substituting $y_t = (\ab_{I_t}^\top\btheta_t + \xi_t) + c_t$ into the reward estimator $\gh_t(i) = \kt_t(i, I_t)\, y_t + 4\eta v_{ti}$ decomposes $\gh_t(i)$ into its clean part $\gh^0_t(i)$, its corruption shift, and the variance bonus:
\begin{align}
  \gh_t(i) = \gh^0_t(i) + c_t\,\kt_t(i, I_t) + 4\eta v_{ti}, \qquad\text{where}\quad \gh^0_t(i) \coloneqq \kt_t(i, I_t)\big(\ab_{I_t}^\top\btheta_t + \xi_t\big). \label{eq:est-split}
\end{align}

The smoothing error $T_0$ is bounded deterministically by $1$: since $|\ab_i^\top \btheta_t| \le 1$ (Assumption~\ref{ass:actions}) and $\sum_{i=1}^K \rho_i = 1$, each round contributes
\begin{align}\label{eq:T0-bound}
  \ab_{i^\star}^\top\btheta_t - \sum_{i=1}^K u_i \ab_i^\top\btheta_t
  = \frac{1}{2T}\Big(\ab_{i^\star}^\top\btheta_t - \sum_{i=1}^K \rho_i \ab_i^\top\btheta_t\Big)
  \le \frac{1}{2T}(1 + 1) = \frac{1}{T},
  \qquad\text{so}\qquad T_0 \le 1 .
\end{align}
We now bound the FTRL regret $T_1$ (Section~\ref{sec:tool-ftrl}) and the estimation errors $T_2$ and $T_3$ (Section~\ref{sec:tool-conc}), and then combine the four bounds in Section~\ref{sec:proof-complete}.

\subsection{Bounding the FTRL Regret}\label{sec:tool-ftrl}
To bound the FTRL term $T_1 = \sum_{t=1}^T \inner{\ghb_t}{\ub - \pb_t}$, we first express the update in line~3 of Algorithm~\ref{alg:main} as a standard FTRL step with a time-varying regularizer. Once $\Qb_t$ is fixed at its inner minimizer $\Qb_t = \argmin_{\Qb \succeq \Qb_{t-1}} J(\Qb, \pb_t)$, the outer iterate $\pb_t$ maximizes the round-$t$ objective $\Lambda_t(\pb) \coloneqq \inner{\Sbb_{t-1}}{\pb} - \psi_t(\pb)$ over $\Pg$, where
\begin{align}\label{eq:psi-t-def}
  \psi_t(\pb) \coloneqq -\sum_{i=1}^K r_{ti}\, h(p_i) = \sum_{i=1}^K r_{ti}\big(\eta^{-1} p_i \log p_i - \alpha p_i\big)
\end{align}
is the radius-weighted regularizer at round $t$ (Lemma~\ref{lem:frozen} in Appendix~\ref{app:lemmas-main}). Setting $\psi_{T+1} \coloneqq \psi_T$ and letting $\pb_{T+1} \in \argmax_{\pb \in \Pg}\Lambda_{T+1}(\pb)$ for the analysis, a standard telescoping FTRL inequality for time-varying regularizers \citep[Exercise~28.12]{lattimore2020bandit} (stated as Lemma~\ref{lem:ftrl-general} in Appendix~\ref{app:lemmas-main}) decomposes $T_1$ into three parts:
\begin{align}
  T_1 = \sum_{t=1}^T \inner{\ghb_t}{\ub - \pb_t}
  &\le \underbrace{\psi_T(\ub)}_{\text{comparator penalty}}
  + \underbrace{\Big(-\psi_1(\pb_1) + \sum_{t=1}^{T-1} \big[\psi_t(\pb_{t+1}) - \psi_{t+1}(\pb_{t+1})\big]\Big)}_{\text{initial and regularizer drift terms}} \nonumber\\
  &\quad + \underbrace{\sum_{t=1}^T \big[\Lambda_t(\pb_{t+1}) - \Lambda_t(\pb_t) + \inner{\ghb_t}{\pb_{t+1} - \pb_t}\big]}_{\text{stability terms}} . \label{eq:ftrl-decomp}
\end{align}
We bound the three parts on the right-hand side of~\eqref{eq:ftrl-decomp} in turn (full proofs of the supporting lemmas are given in Appendix~\ref{app:lemmas-main}).

\paragraph{Comparator penalty.}
Evaluating the final regularizer $\psi_T$ at the comparator $\ub \in \Pg$ gives
\begin{align*}
  \psi_T(\ub) = \sum_{i=1}^K r_{Ti}\big(\eta^{-1} u_i \log u_i - \alpha u_i\big) \le -\alpha\sum_{i=1}^K u_i r_{Ti} = -\alpha\inner{\ub}{\rb_T} ,
\end{align*}
where the inequality uses $u_i \log u_i \le 0$ for $u_i \in (0,1]$ and $r_{Ti} > 0$. This negative linear penalty in the final radii $\rb_T$ is what will later cancel the comparator's estimation and corruption errors in $T_2$.

\paragraph{Initial and regularizer drift terms.}
Because $J(\Qb, \pb) = -\psi_{\Qb}(\pb) - \frac{\kappa}{2}\log\det \Qb$, the round-$t$ regularizer satisfies $-\psi_t(\pb) = J(\Qb_t, \pb) + \frac{\kappa}{2}\log\det \Qb_t$. Since $\Qb_{t+1}$ minimizes $J(\Qb, \pb_{t+1})$ over $\Qb \succeq \Qb_t$ and $\Qb_t$ is feasible, we have $J(\Qb_{t+1}, \pb_{t+1}) \le J(\Qb_t, \pb_{t+1})$, which immediately bounds each round's regularizer drift by the log-determinant increment:
\begin{align*}
  \psi_t(\pb_{t+1}) - \psi_{t+1}(\pb_{t+1})
  = J(\Qb_{t+1}, \pb_{t+1}) - J(\Qb_t, \pb_{t+1}) + \frac{\kappa}{2}\log\frac{\det \Qb_{t+1}}{\det \Qb_t}
  \le \frac{\kappa}{2}\log\frac{\det \Qb_{t+1}}{\det \Qb_t} .
\end{align*}
Likewise, at $t = 1$, $J(\Qb_1, \pb_1) \le J(\Qb_0, \pb_1) = \sum_{i=1}^K h(p_{1i})\norm{\xb_i}$ (since $\Qb_0 = \Ib$). Because every $\pb \in \Pg$ satisfies $p_i \ge \rho_i/(2T) \ge 1/(4KT)$ (Lemma~\ref{lem:design}(a)), the weight $h(p_i) = p_i(\alpha + \eta^{-1}\log(1/p_i))$ is sandwiched between linear multiples of $p_i$ (Lemma~\ref{lem:sandwich}):
\begin{align*}
  \alpha p_i \le h(p_i) \le \kappa p_i, \qquad \text{where } \kappa = \alpha + \eta^{-1}\log(4KT) \le 2\alpha .
\end{align*}
Combined with $\norm{\xb_i} \le \sqrt{2(d+1)}$ (Lemma~\ref{lem:design}(c)), this bounds the initial term by $-\psi_1(\pb_1) = J(\Qb_1, \pb_1) + \frac{\kappa}{2}\log\det \Qb_1 \le \kappa\sqrt{2(d+1)} + \frac{\kappa}{2}\log\det \Qb_1$. Telescoping the log-determinant ratios over $t = 1, \dots, T-1$ and using the eigenvalue growth bound $\log\det \Qb_T \le 2(d+1)\log\big(1 + 4\sqrt2\,(d+1)^{3/2}T^2\big)$ from the first-order optimality conditions of $\Qb_t$ (Lemma~\ref{lem:envelope}(iii)) shows that the initial and drift terms sum to at most
\begin{align*}
  -\psi_1(\pb_1) + \sum_{t=1}^{T-1} \big[\psi_t(\pb_{t+1}) - \psi_{t+1}(\pb_{t+1})\big]
  \le \kappa\Big(\sqrt{2(d+1)} + \frac12\log\det \Qb_T\Big) \le \kappa A_T .
\end{align*}

\paragraph{Stability terms.}
With the radii frozen at $\rb_t$ in round $t$, $\psi_t$ is a radius-weighted negative entropy. By local strong convexity of weighted entropy together with the magnitude and second-moment bounds on the reward estimate $\ghb_t$, each round's stability term is bounded as follows.

\begin{lemma}[{Per-Round Stability}]\label{lem:stability-round}
For every $t \in [T]$ and every $\qb \in \Pg$,
\begin{align*}
  \Lambda_t(\qb) - \Lambda_t(\pb_t) + \inner{\ghb_t}{\qb - \pb_t}
  \le \eta \sum_{i=1}^K p_{ti}\frac{\gh_t(i)^2}{r_{ti}}
  \le 40\eta .
\end{align*}
\end{lemma}
\noindent\emph{Proof.} See Appendix~\ref{app:lemmas-main}.

Applying Lemma~\ref{lem:stability-round} with $\qb = \pb_{t+1}$ and summing over $t = 1, \dots, T$ bounds the stability terms in~\eqref{eq:ftrl-decomp} by $40\eta T$.

Combining the bounds on the comparator penalty, the regularizer drift, and the stability terms in~\eqref{eq:ftrl-decomp} yields the following deterministic bound on $T_1$.

\begin{lemma}[{FTRL with Growing Radii}]\label{lem:ftrl}
On every sample path and for every $\ub \in \Pg$,
\begin{align*}
  T_1 = \sum_{t=1}^T \inner{\ghb_t}{\ub - \pb_t} \le -\alpha\inner{\ub}{\rb_T} + \kappa A_T + 40\eta T .
\end{align*}
\end{lemma}
\noindent\emph{Proof.} See Appendix~\ref{app:lemmas-main}.

\subsection{Bounding the Comparator and Learner Errors}\label{sec:tool-conc}
We next bound the comparator error $T_2 = \sum_{t=1}^T (\sum_{i=1}^K u_i \ab_i^\top\btheta_t - \inner{\ub}{\ghb_t})$ and the learner error $T_3 = \sum_{t=1}^T (\inner{\pb_t}{\ghb_t} - \ab_{I_t}^\top\btheta_t)$ using the estimator decomposition~\eqref{eq:est-split} and the algebraic and moment properties of the normalized kernel $\kt_t$ and variance bonus $\vb_t$ derived in Sections~\ref{sec:motivation} and~\ref{sec:ftrl-algo} (summarized formally in Lemma~\ref{lem:kernel} in Appendix~\ref{app:lemmas-main}).

\paragraph{Comparator error $T_2$.}
Substituting the estimator split $\gh_t(i) = \gh^0_t(i) + c_t\,\kt_t(i, I_t) + 4\eta v_{ti}$ from~\eqref{eq:est-split} into $T_2$ separates the comparator error into a clean estimation error and a corruption shift:
\begin{align}\label{eq:T2-split}
  T_2 = \underbrace{\sum_{i=1}^K u_i\sum_{t=1}^T\big[\ab_i^\top\btheta_t - \gh^0_t(i) - 4\eta v_{ti}\big]}_{\text{clean estimation error}} - \underbrace{\sum_{t=1}^T c_t\sum_{i=1}^K u_i\,\kt_t(i, I_t)}_{\text{corruption shift}} .
\end{align}
For the clean estimation error, because $\kt_t$ reproduces the whitened actions ($\sum_{j=1}^K p_{tj}\kt_t(i,j)\, \xb_j = \xb_i$, Lemma~\ref{lem:kernel}(a)) and the clean rewards are linear in $\xb_j$, the clean part $\gh^0_t(i) = \kt_t(i, I_t)(\ab_{I_t}^\top\btheta_t + \xi_t)$ satisfies $\EE[\gh^0_t(i) \mid \cF_{t-1}] = \ab_i^\top \btheta_t$. Dividing the zero-mean error $\ab_i^\top \btheta_t - \gh^0_t(i)$ by the predictable radius $r_{ti}$ produces a martingale difference $X_t$ with $|X_t| \le 7$ and conditional variance at most $\sigma_t^2 \coloneqq 4v_{ti}/r_{ti}$. Applying a Freedman-style exponential supermartingale bound uniformly over all suffix sums $\sum_{t=\tau}^T (X_t - \eta\sigma_t^2)$ and summing by parts against the nondecreasing sequence $r_{1i} \le \dots \le r_{Ti}$ yields the following high-probability concentration bound.

\begin{lemma}[{Suffix Concentration}]\label{lem:concentration}
Let $L = \log(8KT/\delta)$. With probability at least $1 - \delta/2$, simultaneously for all $i \in [K]$,
\begin{align*}
  \sum_{t=1}^T \big[\ab_i^\top\btheta_t - \gh^0_t(i) - 4\eta v_{ti}\big] \le \frac{L}{\eta}\, r_{Ti}.
\end{align*}
\end{lemma}
\noindent\emph{Proof.} See Appendix~\ref{app:lemmas-main}.

For the corruption shift in~\eqref{eq:T2-split}, the optimality of $\Qb_t$ ensures $\xb_i^\top \Gb_t^{-1} \xb_i \le r_{ti}^2$ (Lemma~\ref{lem:envelope}(i)), which yields the magnitude bound $|\kt_t(i, I_t)| \le 3r_{ti}$ (Lemma~\ref{lem:kernel}(c)). Combining Lemma~\ref{lem:concentration} with $|\kt_t(i, I_t)| \le 3r_{ti}$ and the monotonicity of the radii $r_{ti} \le r_{Ti}$ (Lemma~\ref{lem:envelope}(ii)) bounds $T_2$ on the $(1-\delta/2)$-probability event of Lemma~\ref{lem:concentration}:
\begin{align}
  T_2 &\lstep{a} \frac{L}{\eta}\inner{\ub}{\rb_T} + \sum_{t=1}^T \abr{c_t} \sum_{i=1}^K u_i\, \abr{\kt_t(i, I_t)}
  \lstep{b} \frac{L}{\eta}\inner{\ub}{\rb_T} + 3\sum_{t=1}^T \abr{c_t} \inner{\ub}{\rb_t} \notag\\
  &\lstep{c} \frac{L}{\eta}\inner{\ub}{\rb_T} + 3\inner{\ub}{\rb_T}\sum_{t=1}^T \abr{c_t}
  \lstep{d} \Big(\frac{L}{\eta} + 3C\Big)\inner{\ub}{\rb_T} , \label{eq:T2-bound}
\end{align}
where (a) applies Lemma~\ref{lem:concentration} and the triangle inequality, (b) uses $\abr{\kt_t(i, I_t)} \le 3r_{ti}$, (c) uses $r_{ti} \le r_{Ti}$ and $u_i \ge 0$ for all $i \in [K]$ (so $\inner{\ub}{\rb_t} = \sum_{i=1}^K u_i r_{ti} \le \sum_{i=1}^K u_i r_{Ti} = \inner{\ub}{\rb_T}$), and (d) uses $\sum_{t=1}^T \abr{c_t} \le C$ (Assumption~\ref{ass:protocol}(c)).

\paragraph{Learner error $T_3$.}
On the learner side, the rank-one normalization in~\eqref{eq:kt-vt} guarantees the unit column-sum property $\sum_{i=1}^K p_{ti}\kt_t(i, I_t) = 1$ (Lemma~\ref{lem:kernel}(b)). Taking the inner product of $\pb_t$ with $\gh_t(i) = \kt_t(i, I_t)\, y_t + 4\eta v_{ti}$ therefore collapses the inverse-propensity estimator back to the observed reward $y_t = \ab_{I_t}^\top\btheta_t + \xi_t + c_t$:
\begin{align*}
  \inner{\pb_t}{\ghb_t} = \bigg(\sum_{i=1}^K p_{ti}\kt_t(i, I_t)\bigg)\big(\ab_{I_t}^\top\btheta_t + \xi_t + c_t\big) + 4\eta\inner{\pb_t}{\vb_t}
  = \ab_{I_t}^\top\btheta_t + \xi_t + c_t + 4\eta\inner{\pb_t}{\vb_t} .
\end{align*}
The clean reward $\ab_{I_t}^\top \btheta_t$ cancels inside $T_3 = \sum_{t=1}^T (\inner{\pb_t}{\ghb_t} - \ab_{I_t}^\top \btheta_t)$, leaving
\begin{align}
  T_3 = \sum_{t=1}^T\big(\xi_t + c_t + 4\eta\inner{\pb_t}{\vb_t}\big)
  \lstep{a} \sum_{t=1}^T \xi_t + C + 8\eta T
  \lstep{b} \sqrt{2T\log(2/\delta)} + C + 8\eta T , \label{eq:T3-bound}
\end{align}
where (a) uses $\sum_{t=1}^T c_t \le C$ and $\inner{\pb_t}{\vb_t} \le 2$ (Lemma~\ref{lem:kernel}(e)), and (b) holds with probability at least $1-\delta/2$ by the Azuma--Hoeffding inequality since $(\xi_t)_{t=1}^T$ is a martingale difference sequence bounded in $[-1,1]$ (Lemma~\ref{lem:azuma} in Appendix~\ref{app:lemmas-main}).

Combining~\eqref{eq:T2-bound} and~\eqref{eq:T3-bound} by a union bound over the two $(1-\delta/2)$-probability events yields the following high-probability guarantee for $T_2$ and $T_3$.

\begin{lemma}[{Comparator and Learner Errors}]\label{lem:est-errors}
Let $L = \log(8KT/\delta)$. With probability at least $1 - \delta$, simultaneously for every comparator $\ub \in \Pg$,
\begin{align*}
  T_2 &\le \Big(\frac{L}{\eta} + 3C\Big)\inner{\ub}{\rb_T}, \\
  T_3 &\le \sqrt{2T\log(2/\delta)} + C + 8\eta T .
\end{align*}
\end{lemma}
\noindent\emph{Proof.} See Appendix~\ref{app:lemmas-main}.

\subsection{Combining the Four Bounds}\label{sec:proof-complete}
We now combine the bounds on $T_0, T_1, T_2, T_3$ to obtain Theorem~\ref{thm:main} (formalized in Appendix~\ref{app:app-main}). On the $(1 - \delta)$-probability event of Lemma~\ref{lem:est-errors} (which does not depend on the choice of $i^\star \in [K]$), summing $T_0 \le 1$ from~\eqref{eq:T0-bound}, Lemma~\ref{lem:ftrl} for $T_1$, and Lemma~\ref{lem:est-errors} for $T_2$ and $T_3$ gives
\begin{align*}
  \sum_{t=1}^T (\ab_{i^\star} - \ab_{I_t})^\top\btheta_t
  &\le 1 + \Big(-\alpha + \frac{L}{\eta} + 3C\Big)\inner{\ub}{\rb_T} + \kappa A_T + 48\eta T + C + \sqrt{2T\log(2/\delta)} \\
  &\le 3A_T C + \frac{2A_T L}{\eta} + 48\eta T + C + \sqrt{2T\log(2/\delta)} + 1 ,
\end{align*}
where the choice $\alpha = 3C + L/\eta$ cancels the comparator-dependent term $\inner{\ub}{\rb_T}$ identically and $\kappa = \alpha + \log(4KT)/\eta \le 3C + 2L/\eta$. Taking the maximum over $i^\star \in [K]$ yields Theorem~\ref{thm:main}.

\section{Conclusion and Future Work}\label{sec:conclusion}
We studied adversarial linear bandits with strong corruption and proposed CRW-FTRL, an FTRL algorithm with a radius-weighted entropy regularizer and normalized kernel reward estimates. Its key ingredient is a set of nondecreasing per-action radii, chosen jointly with the sampling distribution by a monotone log-determinant barrier problem, which bound the influence of a corrupted observation on each reward estimate regardless of which action was drawn. With these radii, a negative linear term in the regularizer cancels the corruption bias and the estimation error of every comparator. As a result, CRW-FTRL achieves $\tilde O(\sqrt{dT\log K} + dC)$ regret with high probability against an adaptive adversary, and its application to a discretization of a general compact action set yields $\tilde O(d\sqrt{T} + dC)$ regret, matching the lower bounds up to logarithmic factors and answering the open question of \citet{liu2024corruption} on the statistical rate.

Several directions remain open. First, our discretization argument is not computationally efficient for continuous action sets, and achieving $\tilde O(d\sqrt{T} + dC)$ regret with a polynomial-time algorithm remains open. Even for finite action sets, each round of CRW-FTRL solves a max-min problem whose inner objective is only geodesically convex, and designing efficient solvers for it with provable guarantees is an interesting practical question. Second, like most algorithms for strong corruption \citep{bogunovic2021stochastic,he2022nearly,liu2024corruption}, CRW-FTRL requires the corruption budget $C$ as an input, and it would be interesting to understand the best achievable regret when $C$ is unknown, where \citet{he2022nearly} show that a trade-off with the corruption-free regret is unavoidable in the stochastic setting. Finally, extending our radius-based estimators to linear contextual bandits with changing action sets and to reinforcement learning with linear function approximation is a promising direction.

\section*{AI Usage Statement}
Large language models, specifically Gemini 4 Argon and Claude Opus 5.5, played a central role in this work. Beyond assisting with the writing and editing of the manuscript, the models proposed the core algorithm design and the main proof strategy, and carried out the detailed mathematical derivations and technical verification. The authors formulated the research problem, guided the iterative development of the ideas, audited the proofs, and take full responsibility for all results and claims in the paper.

\appendix
\section{Proofs of the Main Results}\label{app:app-main}

\subsection{Proof of Theorem~\ref{thm:main}}
\begin{proof}[Proof of Theorem~\ref{thm:main}]
Fix an arbitrary comparator action $i^\star \in [K]$ and let $\ub = (1 - \frac{1}{2T})\eb_{i^\star} + \frac{1}{2T}\brho$. Since $T \ge 1$ and $\eb_{i^\star}, \brho \in \Delta_K$, we have $\ub \in \Delta_K$ and $u_i \ge \rho_i/(2T)$ for all $i \in [K]$, so $\ub \in \Pg$. By the decomposition~\eqref{eq:decomp} and the smoothing bound~\eqref{eq:T0-bound} in Section~\ref{sec:barrier},
\begin{align*}
  \sum_{t=1}^T (\ab_{i^\star} - \ab_{I_t})^\top\btheta_t = T_0 + T_1 + T_2 + T_3 \qquad \text{with } T_0 \le 1 .
\end{align*}
Let $\cE$ be the event of Lemma~\ref{lem:est-errors}, which holds with probability at least $1-\delta$ and does not depend on $i^\star$. On $\cE$, applying Lemma~\ref{lem:ftrl} to $T_1$ and Lemma~\ref{lem:est-errors} to $T_2$ and $T_3$ with $\ub \in \Pg$ gives
\begin{align*}
  \sum_{t=1}^T (\ab_{i^\star} - \ab_{I_t})^\top\btheta_t
  &\lstep{a} 1 + \Big(-\alpha\inner{\ub}{\rb_T} + \kappa A_T + 40\eta T\Big) + \Big(\frac{L}{\eta} + 3C\Big)\inner{\ub}{\rb_T} \\
  &\qquad + \Big(\sqrt{2T\log(2/\delta)} + C + 8\eta T\Big) \\
  &\estep{b} \Big(-\alpha + \frac{L}{\eta} + 3C\Big)\inner{\ub}{\rb_T} + \kappa A_T + 48\eta T + C + \sqrt{2T\log(2/\delta)} + 1 \\
  &\estep{c} \kappa A_T + 48\eta T + C + \sqrt{2T\log(2/\delta)} + 1 \\
  &\lstep{d} 3A_T C + \frac{2A_T L}{\eta} + 48\eta T + C + \sqrt{2T\log(2/\delta)} + 1 ,
\end{align*}
where (a) substitutes the four bounds; (b) groups the $\inner{\ub}{\rb_T}$ and $\eta T$ terms; (c) uses $\alpha = 3C + L/\eta$; and (d) uses $\kappa = \alpha + \log(4KT)/\eta \le 3C + 2L/\eta$ (since $\log(4KT) < \log(8KT/\delta) = L$ for $\delta \in (0,1)$) and $A_T \ge 0$. Taking the maximum over $i^\star \in [K]$ on $\cE$ completes the proof.
\end{proof}

\subsection{Proof of Corollary~\ref{cor:rate}}
\begin{proof}[Proof of Corollary~\ref{cor:rate}]
For $\eta = \min\{1/12,\ \sqrt{A_T L/(24T)}\}$, we have $\eta^{-1} = \max\{12,\ \sqrt{24T/(A_T L)}\}$ and $\eta \le \sqrt{A_T L/(24T)}$, so the $\eta$-dependent terms in Theorem~\ref{thm:main} satisfy
\begin{align*}
  \frac{2A_T L}{\eta} + 48\eta T = O\big(A_T L + \sqrt{A_T L T}\big) .
\end{align*}
Since $A_T = \Theta(d\log(dT)) \ge 1$ and $L = \log(8KT/\delta) \ge \max\{\log 8,\, \log(2/\delta)\} \ge 1$, substituting this bound into Theorem~\ref{thm:main} gives, with probability at least $1-\delta$,
\begin{align*}
  \mathrm{Reg}_T = O\big(A_T C + A_T L + \sqrt{A_T L T}\big) .
\end{align*}
Moreover, because $|\ab_i^\top\btheta_t| \le 1$ (Assumption~\ref{ass:actions}), we always have $\mathrm{Reg}_T \le 2T$; thus when $T < A_T L$, $\mathrm{Reg}_T \le 2T < 2\sqrt{A_T L T}$, so the $A_T L$ term is absorbed into $O(\sqrt{A_T L T})$ in all cases. Substituting $A_T = O(d\log(dT))$ and $L = O(\log(KT/\delta))$ yields the stated bound.
\end{proof}

\section{Proofs of the Lemmas in Section~\ref{sec:overview}}\label{app:lemmas-main}
Throughout the appendices, Algorithm~\ref{alg:main} is run with the parameters $\eta \in (0, 1/12]$, $L = \log(8KT/\delta)$, $\alpha = 3C + L/\eta$, and $\kappa = \alpha + \log(4KT)/\eta$ specified in Theorem~\ref{thm:main}. We begin with two foundational lemmas that are used across Sections~\ref{app:sub-ftrl}--\ref{app:sub-est} (all supporting lemmas stated in this section are proved in Appendix~\ref{app:lemmas-B}). First, by construction in Algorithm~\ref{alg:main}, $\pb_t$ maximizes $\inner{\Sbb_{t-1}}{\pb} + \min_{\Qb \succeq \Qb_{t-1}} J(\Qb,\pb)$ jointly with $\Qb_t$. The following lemma shows that all round-$t$ quantities computed before sampling $I_t$ are $\cF_{t-1}$-measurable, and that once $\Qb_t$ is fixed at its inner minimizer, $\pb_t$ is a standard FTRL iterate for the round-$t$ regularizer $\psi_t$.

\begin{lemma}[{Predictability and Frozen Form}]\label{lem:frozen}
For every $t \in [T]$, the quantities $\pb_t$, $\Qb_t$, $\rb_t$, $\Gb_t$, $\kt_t$ and $\vb_t$ are $\cF_{t-1}$-measurable, and $\ghb_t$ and $\Sbb_t$ are $\cF_t$-measurable. Moreover, with the regularizer $\psi_t(\pb) \coloneqq -\sum_i r_{ti}\, h(p_i) = \sum_i r_{ti}\big(\eta^{-1} p_i\log p_i - \alpha p_i\big)$ of round $t$, the iterate has the frozen form
\begin{align}\label{eq:frozen}
  \pb_t = \operatorname*{arg\,max}_{\pb \in \Pg} \big\{\inner{\Sbb_{t-1}}{\pb} - \psi_t(\pb)\big\}.
\end{align}
\end{lemma}

Second, the next lemma summarizes the algebraic and moment properties of the normalized kernel $\kt_t$ and variance bonus $\vb_t$ under the envelope condition $\xb_i^\top \Gb_t^{-1} \xb_i \le r_{ti}^2$.

\begin{lemma}[{Kernel}]\label{lem:kernel}
Let $\pb \in \Pg$ and $r_i \ge 1$, let $\Gb = \sum_{i=1}^K (p_i/r_i)\, \xb_i \xb_i^\top \succ 0$, and suppose that $\xb_i^\top \Gb^{-1} \xb_i \le r_i^2$ for all $i \in [K]$. Then for all $i, j \in [K]$:
\begin{enumerate}
  \item[(a)] $\sum_{j=1}^K p_j \kt(i,j)\, \xb_j = \xb_i$;
  \item[(b)] $\sum_{i=1}^K p_i \kt(i,j) = 1$;
  \item[(c)] $|\kt(i,j)| \le 3 r_i$;
  \item[(d)] $\sum_{i=1}^K (p_i/r_i)\, \kt(i,j)^2 \le 2$;
  \item[(e)] $0 \le v_i \le 9 r_i$ and $\sum_{i=1}^K p_i v_i \le 2$.
\end{enumerate}
\end{lemma}

\subsection{The FTRL Bound}\label{app:sub-ftrl}
To prove the per-round stability bound (Lemma~\ref{lem:stability-round}), we use the following local strong convexity inequality for a radius-weighted negative entropy regularizer with fixed radii $\rb$. By the frozen form~\eqref{eq:frozen}, $\pb_t$ maximizes $\Lambda_t(\pb) = \inner{\Sbb_{t-1}}{\pb} - \psi_t(\pb)$ over $\Pg$, so once we verify the scale condition $\eta\gh_t(i) \le r_{ti}$ using Lemma~\ref{lem:kernel}(c),(e), Lemma~\ref{lem:stability} bounds the round-$t$ stability term by $\eta\sum_i p_{ti}\gh_t(i)^2/r_{ti}$.

\begin{lemma}[{Stability of Weighted Entropy}]\label{lem:stability}
Let $\rb \in \RR^K_{>0}$, $\psi(\pb) = \sum_i r_i\rbr{\eta^{-1}p_i\log p_i - \alpha p_i}$, $\Lambda(\pb) = \inner{\Sbb}{\pb} - \psi(\pb)$ and $\pb^\star \in \argmax_{\Pg}\Lambda$. Let $\gb \in \RR^K$ satisfy $\eta g_i \le r_i$ for all $i$. Then for every $\qb \in \Pg$,
\begin{align*}
  \Lambda(\qb) - \Lambda(\pb^\star) + \inner{\gb}{\qb - \pb^\star} \le \eta\sum_i p^\star_i\, g_i^2 / r_i .
\end{align*}
\end{lemma}

\begin{proof}[Proof of Lemma~\ref{lem:stability-round}]
We first check that the round-$t$ reward estimate satisfies $\eta\,\gh_t(i) \le r_{ti}$ for all $i \in [K]$:
\begin{align*}
  \frac{\eta\,\gh_t(i)}{r_{ti}} \lstep{a} \frac{\eta\big(|\kt_t(i,I_t)|\,|y_t| + 4\eta v_{ti}\big)}{r_{ti}} \lstep{b} \frac{\eta\big(3r_{ti}\cdot 3 + 4\eta \cdot 9 r_{ti}\big)}{r_{ti}} = 9\eta + 36\eta^2 \lstep{c} 1,
\end{align*}
where (a) uses the definition of $\gh_t(i)$, the triangle inequality and $v_{ti} \ge 0$; (b) uses Lemma~\ref{lem:kernel}(c)(e) and $|y_t| \le 3$; and (c) uses $\eta \le 1/12$, which gives $9\eta + 36\eta^2 \le 3/4 + 1/4 = 1$. By the frozen form~\eqref{eq:frozen}, $\pb_t \in \argmax_{\Pg} \Lambda_t$, so Lemma~\ref{lem:stability} applied with $\Sbb = \Sbb_{t-1}$, $\rb = \rb_t$, $\gb = \ghb_t$ and $\pb^\star = \pb_t$ yields the first inequality of Lemma~\ref{lem:stability-round} for every $\qb \in \Pg$. For the second inequality,
\begin{align*}
  \sum_i \frac{p_{ti}\gh_t(i)^2}{r_{ti}}
  &\lstep{a} \sum_i \frac{p_{ti}}{r_{ti}}\Big(2\kt_t(i,I_t)^2 y_t^2 + 2\cdot 16\eta^2 v_{ti}^2\Big)
  \lstep{b} 18\sum_i\frac{p_{ti}}{r_{ti}}\kt_t(i,I_t)^2 + 32\eta^2\sum_i p_{ti}v_{ti}\frac{v_{ti}}{r_{ti}} \\
  &\lstep{c} 18 \cdot 2 + 32\eta^2 \cdot 9 \sum_i p_{ti}v_{ti}
  \lstep{d} 36 + 576\eta^2 \lstep{e} 40 ,
\end{align*}
where (a) uses $\gh_t(i) = \kt_t(i,I_t)\, y_t + 4\eta v_{ti}$ and $(a+b)^2 \le 2a^2 + 2b^2$; (b) uses $y_t^2 \le 9$; (c) uses Lemma~\ref{lem:kernel}(d) with $j = I_t$ and $v_{ti}/r_{ti} \le 9$ (Lemma~\ref{lem:kernel}(e)); (d) uses $\sum_i p_{ti}v_{ti} \le 2$ (Lemma~\ref{lem:kernel}(e)); and (e) uses $\eta \le 1/12$. Multiplying by $\eta$ gives the second inequality of Lemma~\ref{lem:stability-round}.
\end{proof}

We now turn to the cumulative FTRL bound (Lemma~\ref{lem:ftrl}). First, the following standard FTRL inequality for time-varying regularizers \citep[Exercise~28.12]{lattimore2020bandit} decomposes $\sum_{t=1}^T \inner{\ghb_t}{\ub - \pb_t}$ into a comparator penalty $\psi_{T+1}(\ub)$, an initial term $-\psi_1(\pb_1)$, regularizer drift terms $\psi_t(\pb_{t+1}) - \psi_{t+1}(\pb_{t+1})$, and the per-round stability terms bounded in Lemma~\ref{lem:stability-round}.

\begin{lemma}[{General FTRL Inequality; cf.\ \citealt[Exercise~28.12]{lattimore2020bandit}}]\label{lem:ftrl-general}
Let $\psi_1, \dots, \psi_{T+1} \colon \Pg \to \RR$ and $\gb_1, \dots, \gb_T \in \RR^K$. Define $\Sbb_t = \sum_{s \le t} \gb_s$ with $\Sbb_0 = \mathbf 0$, $\Lambda_t(\pb) = \inner{\Sbb_{t-1}}{\pb} - \psi_t(\pb)$, and $\pb_t \in \argmax_{\Pg} \Lambda_t$ for $t \in [T+1]$. Then for every $\ub \in \Pg$,
\begin{align*}
  \sum_{t=1}^T \inner{\gb_t}{\ub - \pb_t} \le{}& \psi_{T+1}(\ub) - \psi_1(\pb_1)
  + \sum_{t=1}^T \big[\psi_t(\pb_{t+1}) - \psi_{t+1}(\pb_{t+1})\big]\\
  &+ \sum_{t=1}^T \big[\Lambda_t(\pb_{t+1}) - \Lambda_t(\pb_t) + \inner{\gb_t}{\pb_{t+1} - \pb_t}\big].
\end{align*}
\end{lemma}

To bound the initial term $-\psi_1(\pb_1)$ and the regularizer drift terms $\psi_t(\pb_{t+1}) - \psi_{t+1}(\pb_{t+1})$, we use the optimality of $\Qb_t = \argmin_{\Qb \succeq \Qb_{t-1}} J(\Qb, \pb_t)$ together with three supporting lemmas:
\begin{itemize}[leftmargin=*]
  \item Lemma~\ref{lem:sandwich} sandwiches the regularizer weight $h(p_i)$ between $\alpha p_i$ and $\kappa p_i$;
  \item Lemma~\ref{lem:design} establishes the properties of the reference design $\brho$ and whitened actions $\xb_i$; in particular, item~(c) bounds the initial radii $r_{0i} = \norm{\xb_i} \le \sqrt{2(d+1)}$, while item~(d) expresses the clean rewards $\ab_i^\top \btheta_t$ as linear functions of $\xb_i$;
  \item Lemma~\ref{lem:envelope} establishes the properties of the radii $r_{ti} = r_i(\Qb_t)$ from the first-order optimality of $\Qb_t$: item~(i) guarantees the envelope condition $\xb_i^\top \Gb_t^{-1} \xb_i \le r_{ti}^2$ needed to invoke Lemma~\ref{lem:kernel}; item~(ii) ensures that the radii are at least $1$ and nondecreasing ($r_{t-1,i} \le r_{ti}$); and item~(iii) bounds the final log-determinant $\log\det \Qb_T$ to yield the $\kappa A_T$ drift bound.
\end{itemize}

\begin{lemma}[{Sandwich}]\label{lem:sandwich}
For every $\pb \in \Pg$ and every $i \in [K]$, $\alpha p_i \le h(p_i) \le \kappa p_i$, where
\begin{align}\label{eq:kappa}
  \kappa \le 2\alpha .
\end{align}
\end{lemma}

\begin{lemma}[{Reference Design}]\label{lem:design}
$\Hb(\brho) \succ 0$, and
\begin{enumerate}
  \item[(a)] $\rho_i \ge 1/(2K)$;
  \item[(b)] $\sum_i \rho_i \xb_i \xb_i^\top = \Ib$;
  \item[(c)] $1 \le \|\xb_i\|^2 \le 2(d+1)$;
  \item[(d)] $\ab_i^\top\btheta_t = \xb_i^\top \Hb(\brho)^{1/2}(\btheta_t^\top, 0)^\top$ for all $i$ and $t$.
\end{enumerate}
\end{lemma}

\begin{lemma}[{Envelope}]\label{lem:envelope}
For every $t \in [T]$:
\begin{enumerate}
  \item[(i)] $\Gb_t \succeq \Qb_t^{-1}$, hence $\xb_i^\top \Gb_t^{-1} \xb_i \le r_{ti}^2$ for all $i \in [K]$;
  \item[(ii)] $\Ib = \Qb_0 \preceq \Qb_{t-1} \preceq \Qb_t$ and $1 \le \norm{\xb_i} \le r_{t-1,i} \le r_{ti}$;
  \item[(iii)] $\log\det \Qb_T \le 2(d+1)\log\big(1 + 4\sqrt2\, (d+1)^{3/2} T^2\big)$, hence $\sqrt{2(d+1)} + \tfrac12\log\det \Qb_T \le A_T$.
\end{enumerate}
\end{lemma}

\begin{proof}[Proof of Lemma~\ref{lem:ftrl}]
By the frozen form~\eqref{eq:frozen}, for $t \le T$ the iterate $\pb_t$ maximizes $\Lambda_t = \inner{\Sbb_{t-1}}{\cdot} - \psi_t$ over $\Pg$. We set $\psi_{T+1} \coloneqq \psi_T$ and let $\pb_{T+1}$ maximize $\inner{\Sbb_T}{\pb} - \psi_T(\pb)$ over $\Pg$; this point is used only in the analysis. Lemma~\ref{lem:ftrl-general}, applied with $\gb_t = \ghb_t$, bounds the left side of the claim by a comparator term, an initial term, drift terms and stability terms, which we bound in this order.

The comparator term reduces to its linear part:
\begin{align*}
  \psi_{T+1}(\ub) =  \sum_i r_{Ti}\big(\eta^{-1}u_i\log u_i - \alpha u_i\big) \le  -\alpha\sum_i r_{Ti}u_i = -\alpha\inner{\ub}{\rb_T},
\end{align*}
where the first equality uses $\psi_{T+1} = \psi_T$ and the definition of $\psi_T$, and the inequality uses $u_i\log u_i \le 0$ for $u_i \in (0,1]$ and $r_{Ti} > 0$.

For the initial and drift terms, recall that $J(\Qb, \pb) = -\psi_{\Qb}(\pb) - \frac{\kappa}{2}\log\det \Qb$, so $-\psi_t(\pb) = J(\Qb_t, \pb) + \frac{\kappa}{2}\log\det \Qb_t$ for each $t \in [T]$. Since $\Qb_1$ minimizes $J(\Qb, \pb_1)$ over $\Qb \succeq \Qb_0 = \Ib$, we have $J(\Qb_1, \pb_1) \le J(\Qb_0, \pb_1)$, and therefore
\begin{align*}
  -\psi_1(\pb_1) &\estep{a} J(\Qb_1, \pb_1) + \frac{\kappa}{2}\log\det \Qb_1
  \lstep{b} J(\Qb_0, \pb_1) + \frac{\kappa}{2}\log\det \Qb_1 \\
  &\estep{c} \sum_i h(p_{1i})\norm{\xb_i} + \frac{\kappa}{2}\log\det \Qb_1
  \lstep{d} \kappa\Big(\sqrt{2(d+1)} + \frac12\log\det \Qb_1\Big),
\end{align*}
where (a) uses the definition of $J$; (b) uses $J(\Qb_1, \pb_1) \le J(\Qb_0, \pb_1)$; (c) uses $\Qb_0 = \Ib$, so that $r_i(\Qb_0) = \norm{\xb_i}$ and $\log\det \Qb_0 = 0$; and (d) uses $h(p_{1i}) \le \kappa p_{1i}$ (Lemma~\ref{lem:sandwich}), $\norm{\xb_i} \le \sqrt{2(d+1)}$ (Lemma~\ref{lem:design}(c)), and $\sum_i p_{1i} = 1$. Similarly, for each $t \le T-1$, $\Qb_{t+1}$ minimizes $J(\Qb, \pb_{t+1})$ over $\Qb \succeq \Qb_t$, so $J(\Qb_{t+1}, \pb_{t+1}) \le J(\Qb_t, \pb_{t+1})$ gives
\begin{align*}
  \psi_t(\pb_{t+1}) - \psi_{t+1}(\pb_{t+1})
  = J(\Qb_{t+1}, \pb_{t+1}) - J(\Qb_t, \pb_{t+1}) + \frac{\kappa}{2}\log\frac{\det \Qb_{t+1}}{\det \Qb_t}
  \le \frac{\kappa}{2}\log\frac{\det \Qb_{t+1}}{\det \Qb_t} .
\end{align*}
The drift term with $t = T$ vanishes because $\psi_{T+1} = \psi_T$. Adding the initial term and the drift terms,
\begin{align*}
  -\psi_1(\pb_1) + \sum_{t=1}^{T} \big[\psi_t(\pb_{t+1}) - \psi_{t+1}(\pb_{t+1})\big]
  &\lstep{a} \kappa\Big(\sqrt{2(d+1)} + \frac12\log\det \Qb_1 + \frac12\sum_{t=1}^{T-1}\log\frac{\det \Qb_{t+1}}{\det \Qb_t}\Big) \\
  &\estep{b} \kappa\Big(\sqrt{2(d+1)} + \frac12\log\det \Qb_T\Big) \lstep{c} \kappa A_T,
\end{align*}
where (a) collects the two previous displays and the vanishing term with $t = T$, (b) telescopes the sum, and (c) is Lemma~\ref{lem:envelope}(iii).

Finally, applying Lemma~\ref{lem:stability-round} with $\qb = \pb_{t+1}$ and summing over $t = 1, \dots, T$ bounds the stability terms by $40\eta T$, so
\begin{align*}
  \sum_{t=1}^T \inner{\ghb_t}{\ub - \pb_t} &\le  \psi_{T+1}(\ub) - \psi_1(\pb_1) + \sum_{t=1}^T \big[\psi_t(\pb_{t+1}) - \psi_{t+1}(\pb_{t+1})\big] + 40\eta T \\
  &\le  -\alpha\inner{\ub}{\rb_T} + \kappa A_T + 40\eta T ,
\end{align*}
where the first inequality is Lemma~\ref{lem:ftrl-general} together with Lemma~\ref{lem:stability-round}, and the second uses the bounds on the comparator term and on the initial and drift terms.
\end{proof}

\subsection{The Concentration Bound}\label{app:sub-conc}
We now prove Lemma~\ref{lem:concentration} by combining the predictability of the round-$t$ quantities (Lemma~\ref{lem:frozen}), the linear representation of the clean rewards in the whitened actions (Lemma~\ref{lem:design}(d)), the action reproduction and magnitude bounds of the normalized kernel (Lemma~\ref{lem:kernel}(a),(c)), and the monotonicity of the radii (Lemma~\ref{lem:envelope}(ii)).

\begin{proof}[Proof of Lemma~\ref{lem:concentration}]
Fix $i \in [K]$; at the end we take a union bound over $i$. We normalize each round by its radius and set
\begin{align*}
  X_t = \frac{\ab_i^\top\btheta_t - \gh^0_t(i)}{r_{ti}}, \qquad \sigma_t^2 = \frac{4 v_{ti}}{r_{ti}}, \qquad \bvartheta_t = \Hb(\brho)^{1/2}(\btheta_t^\top, 0)^\top .
\end{align*}
The quantities computed before round $t$, namely $\pb_t$, the kernel $\kt_t$, the radii $r_{ti}$ and the variance proxies $v_{ti}$, are $\cF_{t-1}$-measurable by Lemma~\ref{lem:frozen}. So is $\bvartheta_t$, because $\Hb(\brho)$ is deterministic and $\btheta_t$ is $\cF_{t-1}$-measurable.

We first show that $X_t$ is a martingale difference with bounded range and controlled variance. The mean uses that the clean rewards are linear in the whitened actions, which is Lemma~\ref{lem:design}(d). For the mean,
\begin{align*}
  \EE[\gh^0_t(i) \mid \cF_{t-1}] \estep{a} \sum_j p_{tj}\kt_t(i,j)\,\EE\big[\ab_j^\top\btheta_t + \xi_t \bigm| \cF_{t-1}, I_t = j\big] \estep{b} \sum_j p_{tj}\kt_t(i,j)\, \xb_j^\top\bvartheta_t \estep{c} \xb_i^\top\bvartheta_t \estep{d} \ab_i^\top\btheta_t,
\end{align*}
where (a) conditions on $I_t$, using $\PP(I_t = j \mid \cF_{t-1}) = p_{tj}$ and the $\cF_{t-1}$-measurability of $\kt_t$; (b) uses $\EE[\xi_t \mid \cF_{t-1}, I_t] = 0$ (Assumption~\ref{ass:protocol}(b)) and $\ab_j^\top\btheta_t = \xb_j^\top\bvartheta_t$ (Lemma~\ref{lem:design}(d)); (c) is Lemma~\ref{lem:kernel}(a); and (d) is Lemma~\ref{lem:design}(d). Hence $\EE[X_t \mid \cF_{t-1}] = 0$. For the range,
\begin{align*}
  \abr{X_t} \lstep{a} \frac{\abr{\ab_i^\top\btheta_t} + \abr{\kt_t(i,I_t)}\,\abr{\ab_{I_t}^\top\btheta_t + \xi_t}}{r_{ti}} \lstep{b} \frac{1 + 3r_{ti}\cdot 2}{r_{ti}} \lstep{c} 7,
\end{align*}
where (a) is the triangle inequality; (b) uses $\abr{\ab_i^\top\btheta_t} \le 1$, Lemma~\ref{lem:kernel}(c) and $\abr{\ab_{I_t}^\top\btheta_t + \xi_t} \le 2$; and (c) uses $r_{ti} \ge 1$ (Lemma~\ref{lem:envelope}(ii)). For the variance,
\begin{align*}
  \EE[X_t^2 \mid \cF_{t-1}] \estep{a} \frac{\mathrm{Var}\big(\gh^0_t(i) \mid \cF_{t-1}\big)}{r_{ti}^2} \lstep{b} \frac{\EE[\gh^0_t(i)^2 \mid \cF_{t-1}]}{r_{ti}^2} \lstep{c} \frac{4\sum_j p_{tj}\kt_t(i,j)^2}{r_{ti}^2} \estep{d} \frac{4v_{ti}}{r_{ti}} = \sigma_t^2,
\end{align*}
where (a) uses that $\ab_i^\top\btheta_t$ is the conditional mean of $\gh^0_t(i)$ and that $r_{ti}$ is $\cF_{t-1}$-measurable; (b) bounds the variance by the second moment; (c) conditions on $I_t$ and uses $(\ab_j^\top\btheta_t + \xi_t)^2 \le 4$; and (d) is the definition $v_{ti} = r_{ti}^{-1}\sum_j p_{tj}\kt_t(i,j)^2$.

These properties yield an exponential supermartingale. Since $\eta X_t \le 7\eta \le 7/12 \le 1$,
\begin{align*}
  \EE[e^{\eta X_t} \mid \cF_{t-1}] \lstep{a} 1 + \eta\,\EE[X_t \mid \cF_{t-1}] + \eta^2\,\EE[X_t^2 \mid \cF_{t-1}] \estep{b} 1 + \eta^2\,\EE[X_t^2 \mid \cF_{t-1}] \lstep{c} 1 + \eta^2\sigma_t^2 \lstep{d} e^{\eta^2\sigma_t^2},
\end{align*}
where (a) uses $e^x \le 1 + x + x^2$ for $x \le 1$; (b) uses $\EE[X_t \mid \cF_{t-1}] = 0$; (c) is the variance bound; and (d) uses $1 + x \le e^x$. Fix a start $\tau \in [T]$ and let $Z_s = \exp\big(\sum_{t=\tau}^s(\eta X_t - \eta^2\sigma_t^2)\big)$ for $s \ge \tau - 1$, so that $Z_{\tau-1} = 1$. Since $\sigma_s^2$ is $\cF_{s-1}$-measurable, the last display gives $\EE[Z_s \mid \cF_{s-1}] = Z_{s-1}\,\EE[e^{\eta X_s} \mid \cF_{s-1}]\, e^{-\eta^2\sigma_s^2} \le Z_{s-1}$, so $(Z_s)$ is a supermartingale and $\EE[Z_T] \le 1$. Therefore
\begin{align*}
  \Pr\Big(\sum_{t=\tau}^T (X_t - \eta\sigma_t^2) \ge \tfrac1\eta\log\tfrac{2KT}{\delta}\Big) \estep{a} \Pr\Big(Z_T \ge \frac{2KT}{\delta}\Big) \lstep{b} \frac{\delta}{2KT}\,\EE[Z_T] \lstep{c} \frac{\delta}{2KT},
\end{align*}
where (a) multiplies the inequality inside the probability by $\eta > 0$ and exponentiates, (b) is Markov's inequality, and (c) uses $\EE[Z_T] \le 1$. A union bound over $(i, \tau) \in [K]\times[T]$ shows that, with probability at least $1 - \delta/2$, every suffix sum $U_\tau = \sum_{t \ge \tau}(X_t - \eta\sigma_t^2)$ satisfies $U_\tau \le L/\eta$, since $\log(2KT/\delta) \le L$.

On this event we return to the scale of the radii by summation by parts. Let $U_{T+1} = 0$, so that $X_t - \eta\sigma_t^2 = U_t - U_{t+1}$. Then
\begin{align*}
  \sum_{t=1}^T r_{ti}(X_t - \eta\sigma_t^2) &\estep{a} r_{1i}U_1 + \sum_{t=2}^T(r_{ti} - r_{t-1,i})U_t \\
  &\lstep{b} \frac{L}{\eta}\Big(r_{1i} + \sum_{t=2}^T(r_{ti} - r_{t-1,i})\Big) \estep{c} \frac{L}{\eta}r_{Ti},
\end{align*}
where (a) substitutes $X_t - \eta\sigma_t^2 = U_t - U_{t+1}$, regroups the terms by $U_t$ and uses $U_{T+1} = 0$; (b) uses $U_t \le L/\eta$ together with $r_{1i} \ge 0$ and $r_{ti} - r_{t-1,i} \ge 0$ (Lemma~\ref{lem:envelope}(ii)); and (c) telescopes the sum. Finally, the definitions of $X_t$ and $\sigma_t^2$ give $r_{ti}(X_t - \eta\sigma_t^2) = \ab_i^\top\btheta_t - \gh^0_t(i) - 4\eta v_{ti}$, so the left side equals $\sum_t[\ab_i^\top\btheta_t - \gh^0_t(i) - 4\eta v_{ti}]$, which proves the lemma for all $i \in [K]$ simultaneously.
\end{proof}

\subsection{The Comparator and Learner Error Bounds}\label{app:sub-est}
To prove Lemma~\ref{lem:est-errors}, we combine Lemma~\ref{lem:concentration}, the algebraic properties of the normalized kernel (Lemma~\ref{lem:kernel}(b),(c),(e)), and the envelope and monotonicity properties of the radii (Lemma~\ref{lem:envelope}(i),(ii)) with the following concentration bound for the cumulative observation noise $\sum_{t=1}^T \xi_t$.

\begin{lemma}[{Noise}]\label{lem:azuma}
With probability at least $1 - \delta/2$, $\sum_{t=1}^T \xi_t \le \sqrt{2T\log(2/\delta)}$.
\end{lemma}

\begin{proof}[Proof of Lemma~\ref{lem:azuma}]
The noise has zero conditional mean given the past:
\begin{align*}
  \EE[\xi_t \mid \cF_{t-1}] =  \EE\big[\EE[\xi_t \mid \cF_{t-1}, I_t] \bigm| \cF_{t-1}\big] =  0,
\end{align*}
where the first equality is the tower property and the second uses $\EE[\xi_t \mid \cF_{t-1}, I_t] = 0$ (Assumption~\ref{ass:protocol}(b)). Since moreover $\abr{\xi_t} \le 1$ and $\xi_t$ is $\cF_t$-measurable, the sequence $(\xi_t)$ is a martingale difference sequence with increments in $[-1,1]$, and the Azuma--Hoeffding inequality (Lemma~\ref{lem:aux-azuma}) with $\delta' = \delta/2$ gives the claim.
\end{proof}

\begin{proof}[Proof of Lemma~\ref{lem:est-errors}]
At every round $t \in [T]$, the iterate $\pb_t \in \Pg$ satisfies $p_{ti} \ge \rho_i/(2T) > 0$ (since $\rho_i \ge 1/(2K)$ by Lemma~\ref{lem:design}(a)), and the radii satisfy $r_{ti} \ge 1$ by Lemma~\ref{lem:envelope}(ii). Because $\{\zb_i\}_{i=1}^K$ spans $\RR^{d+1}$ (Assumption~\ref{ass:actions}) and $\Hb(\brho) \succ 0$ (Lemma~\ref{lem:design}), the whitened actions $\xb_i = \Hb(\brho)^{-1/2}\zb_i$ also span $\RR^{d+1}$, so $\Gb_t = \sum_{i=1}^K (p_{ti}/r_{ti})\, \xb_i \xb_i^\top \succ 0$. Moreover, $\xb_i^\top \Gb_t^{-1} \xb_i \le r_{ti}^2$ holds for all $i \in [K]$ by Lemma~\ref{lem:envelope}(i). Consequently, $(\pb_t, \rb_t, \Gb_t)$ satisfies all hypotheses of Lemma~\ref{lem:kernel} at every round $t \in [T]$.

Let $\cE_{\mathrm{conc}}$ denote the event of Lemma~\ref{lem:concentration}, on which
\begin{align}\label{eq:app-E-conc}
  \sum_{t=1}^T \big[\ab_i^\top\btheta_t - \gh^0_t(i) - 4\eta v_{ti}\big] \le \frac{L}{\eta}\, r_{Ti} \qquad \text{for all } i \in [K] ,
\end{align}
and let $\cE_{\mathrm{noise}}$ denote the event of Lemma~\ref{lem:azuma}, on which
\begin{align}\label{eq:app-E-noise}
  \sum_{t=1}^T \xi_t \le \sqrt{2T\log(2/\delta)} .
\end{align}
By Lemma~\ref{lem:concentration} and Lemma~\ref{lem:azuma}, $\PP(\cE_{\mathrm{conc}}) \ge 1 - \delta/2$ and $\PP(\cE_{\mathrm{noise}}) \ge 1 - \delta/2$. By the union bound, the intersection $\cE \coloneqq \cE_{\mathrm{conc}} \cap \cE_{\mathrm{noise}}$ occurs with probability at least $1 - \delta$. Note that $\cE$ depends only on the algorithm's trajectory and not on the choice of comparator $\ub \in \Pg$.

To bound the comparator error $T_2$, note that by the reward model $y_t = \ab_{I_t}^\top\btheta_t + \xi_t + c_t$ (Assumption~\ref{ass:protocol}(c)) and the definition $\gh_t(i) = \kt_t(i, I_t)\, y_t + 4\eta v_{ti}$, each coordinate of $\ghb_t$ decomposes as
\begin{align}\label{eq:app-est-split}
  \gh_t(i) = \gh^0_t(i) + c_t\,\kt_t(i, I_t) + 4\eta v_{ti}, \qquad\text{where}\quad \gh^0_t(i) = \kt_t(i, I_t)\big(\ab_{I_t}^\top\btheta_t + \xi_t\big) .
\end{align}
Substituting~\eqref{eq:app-est-split} into the definition of $T_2$ gives, for any comparator $\ub \in \Pg$,
\begin{align*}
  T_2 &= \sum_{t=1}^T \Big(\sum_{i=1}^K u_i \ab_i^\top\btheta_t - \inner{\ub}{\ghb_t}\Big) \\
  &= \sum_{i=1}^K u_i \sum_{t=1}^T \big[\ab_i^\top\btheta_t - \gh^0_t(i) - 4\eta v_{ti}\big] - \sum_{t=1}^T c_t \sum_{i=1}^K u_i\,\kt_t(i, I_t) .
\end{align*}
On the event $\cE$, since $u_i \ge 0$ for all $i \in [K]$, multiplying~\eqref{eq:app-E-conc} by $u_i$ and summing over $i \in [K]$ bounds the first sum by $(L/\eta)\sum_{i=1}^K u_i r_{Ti} = (L/\eta)\inner{\ub}{\rb_T}$. For the second sum,
\begin{align*}
  -\sum_{t=1}^T c_t \sum_{i=1}^K u_i\,\kt_t(i, I_t)
  &\lstep{a} \sum_{t=1}^T \abr{c_t} \sum_{i=1}^K u_i\,\abr{\kt_t(i, I_t)}
  \lstep{b} 3\sum_{t=1}^T \abr{c_t} \sum_{i=1}^K u_i r_{ti} \\
  &\lstep{c} 3\sum_{t=1}^T \abr{c_t} \sum_{i=1}^K u_i r_{Ti}
  \estep{d} 3\inner{\ub}{\rb_T}\sum_{t=1}^T \abr{c_t}
  \lstep{e} 3C\inner{\ub}{\rb_T} ,
\end{align*}
where (a) uses the triangle inequality and $u_i \ge 0$; (b) uses $|\kt_t(i, I_t)| \le 3r_{ti}$ from Lemma~\ref{lem:kernel}(c) and $u_i \ge 0$; (c) uses the monotonicity of the radii $r_{ti} \le r_{Ti}$ from Lemma~\ref{lem:envelope}(ii) together with $u_i \ge 0$ and $|c_t| \ge 0$; (d) factors out $\inner{\ub}{\rb_T} = \sum_{i=1}^K u_i r_{Ti}$; and (e) uses $\sum_{t=1}^T |c_t| \le C$ (Assumption~\ref{ass:protocol}(c)) together with $\inner{\ub}{\rb_T} \ge 0$. Combining the two bounds gives $T_2 \le (L/\eta + 3C)\inner{\ub}{\rb_T}$.

Finally, to bound the learner error $T_3$, taking the inner product of $\pb_t$ with $\gh_t(i) = \kt_t(i, I_t)\, y_t + 4\eta v_{ti}$ at each round $t \in [T]$ gives
\begin{align*}
  \inner{\pb_t}{\ghb_t}
  = \bigg(\sum_{i=1}^K p_{ti}\kt_t(i, I_t)\bigg) y_t + 4\eta\inner{\pb_t}{\vb_t}
  \estep{a} y_t + 4\eta\inner{\pb_t}{\vb_t}
  \estep{b} \ab_{I_t}^\top\btheta_t + \xi_t + c_t + 4\eta\inner{\pb_t}{\vb_t} ,
\end{align*}
where (a) uses the unit column-sum property $\sum_{i=1}^K p_{ti}\kt_t(i, j) = 1$ for $j = I_t$ from Lemma~\ref{lem:kernel}(b), and (b) substitutes $y_t = \ab_{I_t}^\top\btheta_t + \xi_t + c_t$. Subtracting $\ab_{I_t}^\top\btheta_t$ and summing over $t = 1, \dots, T$, we obtain on the event $\cE$:
\begin{align*}
  T_3 &= \sum_{t=1}^T \big(\inner{\pb_t}{\ghb_t} - \ab_{I_t}^\top\btheta_t\big)
  = \sum_{t=1}^T \xi_t + \sum_{t=1}^T c_t + 4\eta\sum_{t=1}^T \inner{\pb_t}{\vb_t} \\
  &\lstep{a} \sum_{t=1}^T \xi_t + C + 8\eta T
  \lstep{b} \sqrt{2T\log(2/\delta)} + C + 8\eta T ,
\end{align*}
where (a) uses $\sum_{t=1}^T c_t \le \sum_{t=1}^T |c_t| \le C$ (Assumption~\ref{ass:protocol}(c)) and $\inner{\pb_t}{\vb_t} = \sum_{i=1}^K p_{ti}v_{ti} \le 2$ (Lemma~\ref{lem:kernel}(e)), and (b) uses~\eqref{eq:app-E-noise}. This completes the proof of Lemma~\ref{lem:est-errors}.
\end{proof}

\section{Proofs of the Supporting Lemmas in Appendix~\ref{app:lemmas-main}}\label{app:lemmas-B}

\subsection{FTRL and Regularizer Properties}\label{app:sub-ftrl-props}
To prove Lemma~\ref{lem:frozen}, we use the following two lemmas on the inner minimization problem in line~4 and the outer maximization problem in line~3 of Algorithm~\ref{alg:main} (both proved in Appendix~\ref{app:foundations}): Lemma~\ref{lem:inner} and Lemma~\ref{lem:outer}(a) ensure that $\pb_t$ and $\Qb_t$ are well-defined continuous functions of $(\Sbb_{t-1}, \Qb_{t-1})$ (giving $\cF_{t-1}$-measurability by induction), while Lemma~\ref{lem:outer}(b) shows that $\pb_t$ also maximizes the FTRL objective with $\Qb$ frozen at $\Qb_t$.

\begin{lemma}[{Inner Problem}]\label{lem:inner}
For every $\Db \succeq \Ib$ and $\pb \in \Pg$, the problem $\min_{\Qb \succeq \Db} J(\Qb,\pb)$ has a unique minimizer, denoted $\Qb(\Db,\pb)$. The map $(\Db,\pb) \mapsto \Qb(\Db,\pb)$ is continuous on $\{\Db \succeq \Ib\} \times \Pg$.
\end{lemma}

\begin{lemma}[{Outer Problem}]\label{lem:outer}
Fix $\Sbb \in \RR^K$ and $\Db \succeq \Ib$.
\begin{enumerate}
  \item[(a)] The function $\pb \mapsto \inner{\Sbb}{\pb} + \min_{\Qb \succeq \Db} J(\Qb,\pb)$ has a unique maximizer $\pb(\Sbb,\Db)$ over $\Pg$, and the map $(\Sbb,\Db) \mapsto \pb(\Sbb,\Db)$ is continuous.
  \item[(b)] Let $\pb^\star = \pb(\Sbb,\Db)$ and $\Qb^\star = \Qb(\Db, \pb^\star)$. Then $\pb^\star$ is also the unique maximizer of $\inner{\Sbb}{\pb} + \sum_i r_i(\Qb^\star) h(p_i)$ over $\Pg$.
\end{enumerate}
\end{lemma}

\begin{proof}[Proof of Lemma~\ref{lem:frozen}]
We argue by induction on $t$, starting from the deterministic $\Sbb_0 = \mathbf 0$ and $\Qb_0 = \Ib$. Suppose that $\Sbb_{t-1}$ and $\Qb_{t-1}$ are $\cF_{t-1}$-measurable. By Lemma~\ref{lem:inner} and Lemma~\ref{lem:outer}(a), $\pb_t = \pb(\Sbb_{t-1}, \Qb_{t-1})$ and $\Qb_t = \Qb(\Qb_{t-1}, \pb_t)$ are continuous functions of $(\Sbb_{t-1}, \Qb_{t-1})$, and hence are $\cF_{t-1}$-measurable. The radii $\rb_t$, the matrix $\Gb_t$, the kernel $\kt_t$ and the variance proxies $\vb_t$ are continuous functions of $(\pb_t, \Qb_t)$, so they are $\cF_{t-1}$-measurable as well. The estimate $\ghb_t$ is a function of these quantities and of $(I_t, y_t)$, so it is $\cF_t$-measurable, and so is $\Sbb_t = \Sbb_{t-1} + \ghb_t$, which completes the induction. For the frozen form, we apply Lemma~\ref{lem:outer}(b) with $\Sbb = \Sbb_{t-1}$ and $\Db = \Qb_{t-1}$. Then $\pb^\star = \pb_t$ and $\Qb^\star = \Qb(\Qb_{t-1}, \pb_t) = \Qb_t$, so $\pb_t$ is the unique maximizer over $\Pg$ of $\inner{\Sbb_{t-1}}{\pb} + \sum_i r_{ti}\, h(p_i)$, which equals $\inner{\Sbb_{t-1}}{\pb} - \psi_t(\pb)$ by the definition of $\psi_t$.
\end{proof}

\begin{proof}[Proof of Lemma~\ref{lem:stability}]
The proof bounds the left side by a Bregman divergence and then maximizes coordinatewise. On the positive orthant, $\psi$ is differentiable and convex, with $\partial_i\psi(\pb) = r_i\big(\eta^{-1}(1 + \log p_i) - \alpha\big)$. Its Bregman divergence is
\begin{align*}
  B(\qb, \pb) \coloneqq \psi(\qb) - \psi(\pb) - \inner{\nabla\psi(\pb)}{\qb - \pb} = \eta^{-1}\sum_i r_i\big(q_i\log(q_i/p_i) - q_i + p_i\big),
\end{align*}
where the equality follows by substituting $\partial_i\psi$; the linear term $-\alpha\sum_i r_i p_i$ of $\psi$ cancels. Since $\Lambda$ is a linear function minus $\psi$,
\begin{align*}
  \Lambda(\qb) - \Lambda(\pb^\star) =  \inner{\nabla\Lambda(\pb^\star)}{\qb - \pb^\star} - B(\qb, \pb^\star) \le  -B(\qb,\pb^\star),
\end{align*}
where the equality uses $\Lambda = \inner{\Sbb}{\cdot} - \psi$ and the definition of $B$, and the inequality is the first-order optimality of $\pb^\star$ on the convex set $\Pg$, which gives $\inner{\nabla\Lambda(\pb^\star)}{\qb - \pb^\star} \le 0$. For $s > 0$, let $\varphi_i(s) = g_i(s - p^\star_i) - \frac{r_i}{\eta}\big(s\log(s/p^\star_i) - s + p^\star_i\big)$ and $u_i = \eta g_i/r_i$. Then
\begin{align*}
  \Lambda(\qb) - \Lambda(\pb^\star) + \inner{\gb}{\qb - \pb^\star}
  &\lstep{a} \sum_i \varphi_i(q_i)
  \lstep{b} \sum_i \sup_{s > 0} \varphi_i(s)
  \estep{c} \sum_i \frac{r_i p^\star_i}{\eta}\big(e^{u_i} - 1 - u_i\big) \\
  &\lstep{d} \sum_i \frac{r_i p^\star_i}{\eta}u_i^2
  \estep{e} \eta\sum_i p^\star_i\, \frac{g_i^2}{r_i} ,
\end{align*}
where (a) is the previous display; (b) maximizes each coordinate separately; (c) uses that $\varphi_i$ is concave with $\varphi_i'(s) = g_i - \frac{r_i}{\eta}\log(s/p^\star_i)$, so that its maximum is attained at $s = p^\star_i e^{u_i}$, where $\varphi_i = g_i p^\star_i(e^{u_i} - 1) - \frac{r_i p^\star_i}{\eta}\big(u_i e^{u_i} - e^{u_i} + 1\big) = \frac{r_i p^\star_i}{\eta}(e^{u_i} - 1 - u_i)$ by $g_i = r_i u_i/\eta$; (d) uses $u_i \le 1$, which is the hypothesis $\eta g_i \le r_i$, and $e^u - 1 - u \le u^2$ for $u \le 1$; and (e) substitutes $u_i = \eta g_i/r_i$.
\end{proof}

\begin{proof}[Proof of Lemma~\ref{lem:ftrl-general}]
The proof compares the value of the FTRL objective at the comparator with its values along the iterates. First,
\begin{align*}
  \inner{\Sbb_T}{\ub} =  \Lambda_{T+1}(\ub) + \psi_{T+1}(\ub) \le  \Lambda_{T+1}(\pb_{T+1}) + \psi_{T+1}(\ub),
\end{align*}
where the equality is the definition of $\Lambda_{T+1}$ and the inequality uses that $\pb_{T+1}$ maximizes $\Lambda_{T+1}$ over $\Pg$. To express $\Lambda_{T+1}(\pb_{T+1})$ through the iterates, note that for each $t$,
\begin{align*}
  \Lambda_{t+1}(\pb_{t+1}) &=  \inner{\Sbb_{t-1}}{\pb_{t+1}} + \inner{\gb_t}{\pb_{t+1}} - \psi_{t+1}(\pb_{t+1}) \\
  &=  \Lambda_t(\pb_{t+1}) + \inner{\gb_t}{\pb_{t+1}} + \big[\psi_t(\pb_{t+1}) - \psi_{t+1}(\pb_{t+1})\big],
\end{align*}
where the first equality uses the definition of $\Lambda_{t+1}$ and $\Sbb_t = \Sbb_{t-1} + \gb_t$, and the second adds and subtracts $\psi_t(\pb_{t+1})$ and uses the definition of $\Lambda_t$. Therefore
\begin{align*}
  \Lambda_{T+1}(\pb_{T+1}) &=  \Lambda_1(\pb_1) + \sum_{t=1}^T \big[\Lambda_{t+1}(\pb_{t+1}) - \Lambda_t(\pb_t)\big] \\
  &=  -\psi_1(\pb_1) + \sum_{t=1}^T \big[\Lambda_t(\pb_{t+1}) - \Lambda_t(\pb_t) + \inner{\gb_t}{\pb_{t+1}}\big] + \sum_{t=1}^T \big[\psi_t(\pb_{t+1}) - \psi_{t+1}(\pb_{t+1})\big],
\end{align*}
where the first equality telescopes the sum, and the second uses $\Lambda_1(\pb_1) = -\psi_1(\pb_1)$, which holds because $\Sbb_0 = \mathbf 0$, together with the previous display. Finally,
\begin{align*}
  \sum_{t=1}^T \inner{\gb_t}{\ub - \pb_t} =  \inner{\Sbb_T}{\ub} - \sum_{t=1}^T \inner{\gb_t}{\pb_t} \le  \Lambda_{T+1}(\pb_{T+1}) + \psi_{T+1}(\ub) - \sum_{t=1}^T \inner{\gb_t}{\pb_t},
\end{align*}
where the equality uses $\Sbb_T = \sum_t \gb_t$ and the inequality is the first display. Substituting the expression for $\Lambda_{T+1}(\pb_{T+1})$ and combining $\inner{\gb_t}{\pb_{t+1}} - \inner{\gb_t}{\pb_t} = \inner{\gb_t}{\pb_{t+1} - \pb_t}$ gives the claim.
\end{proof}

\begin{proof}[Proof of Lemma~\ref{lem:sandwich}]
By the definition of $h$, the ratio $h(p_i)/p_i$ is $\alpha$ plus a logarithmic term, which we bound from both sides:
\begin{align*}
  \alpha \lstep{a} \alpha + \frac{1}{\eta}\log\frac{1}{p_i} \estep{b} \frac{h(p_i)}{p_i} \lstep{c} \alpha + \frac{\log(4KT)}{\eta} \estep{d} \kappa ,
\end{align*}
where (a) uses $p_i \le 1$, so that $\log(1/p_i) \ge 0$; (b) is the definition $h(s) = s(\alpha - \eta^{-1}\log s)$; (c) uses $p_i \ge \rho_i/(2T) \ge 1/(4KT)$, which holds by the definition of $\Pg$ and Lemma~\ref{lem:design}(a); and (d) is the definition of $\kappa$. Multiplying by $p_i > 0$ gives $\alpha p_i \le h(p_i) \le \kappa p_i$. Moreover, since $4KT < 8KT/\delta$ implies $\log(4KT) < \log(8KT/\delta) = L$, and $L/\eta \le 3C + L/\eta = \alpha$, we have $\kappa = \alpha + \log(4KT)/\eta \le \alpha + L/\eta \le 2\alpha$.
\end{proof}

\subsection{Design, Radii, and Kernel Properties}\label{app:sub-design-radii-kernel}
\begin{proof}[Proof of Lemma~\ref{lem:kernel}]
Write $\ell_i \coloneqq \sqrt{\xb_i^\top \Gb^{-1} \xb_i} \le r_i$, $W \coloneqq \sum_{m=1}^K p_m r_m$, and $b(j) \coloneqq \sum_{l=1}^K p_l k(l,j)$, so that the normalized kernel in~\eqref{eq:kt-vt} takes the form
\begin{align*}
  \kt(i,j) = k(i,j) - \frac{r_i}{W}\big(b(j) - 1\big) .
\end{align*}
Since $r_m \ge 1$ for all $m \in [K]$ and $\sum_{m=1}^K p_m = 1$, we have $W = \sum_{m=1}^K p_m r_m \ge 1$.

\medskip\noindent\textbf{Preliminary properties of the unnormalized kernel $k$.}
Before proving items (a)--(e), we record four basic properties of $k(i,j) = \xb_i^\top \Gb^{-1} \xb_j / r_j$:
\begin{enumerate}[leftmargin=*]
  \item \emph{Action reproduction:} For every $i \in [K]$,
  \begin{align}\label{eq:k-reprod}
    \sum_{j=1}^K p_j k(i,j) \xb_j^\top = \xb_i^\top \Gb^{-1} \sum_{j=1}^K \frac{p_j}{r_j} \xb_j \xb_j^\top = \xb_i^\top \Gb^{-1} \Gb = \xb_i^\top ,
  \end{align}
  where the first equality uses the definition of $k(i,j)$ and the second uses $\Gb = \sum_{j=1}^K (p_j/r_j)\, \xb_j \xb_j^\top$.
  \item \emph{Pointwise magnitude bound:} For all $i, j \in [K]$,
  \begin{align}\label{eq:k-mag}
    |k(i,j)| \estep{a} \frac{\abr{\xb_i^\top \Gb^{-1} \xb_j}}{r_j} \lstep{b} \frac{\ell_i\ell_j}{r_j} \lstep{c} \ell_i \lstep{d} r_i ,
  \end{align}
  where (a) is the definition of $k$, (b) is the Cauchy--Schwarz inequality for the inner product $(\ub, \vb) \mapsto \ub^\top \Gb^{-1} \vb$, and (c) and (d) use $\ell_j \le r_j$ and $\ell_i \le r_i$.
  \item \emph{Radius-weighted second moment:} For every $j \in [K]$,
  \begin{align}\label{eq:k-2nd}
    \sum_{i=1}^K \frac{p_i}{r_i} k(i,j)^2 \estep{a} \frac{\xb_j^\top \Gb^{-1}\big(\sum_{i=1}^K \frac{p_i}{r_i} \xb_i \xb_i^\top\big) \Gb^{-1} \xb_j}{r_j^2} \estep{b} \frac{\xb_j^\top \Gb^{-1} \xb_j}{r_j^2} = \frac{\ell_j^2}{r_j^2} \lstep{c} 1 ,
  \end{align}
  where (a) writes $k(i,j)^2 = \xb_j^\top \Gb^{-1} \xb_i \xb_i^\top \Gb^{-1} \xb_j / r_j^2$, (b) uses the definition of $\Gb$, and (c) uses $\ell_j \le r_j$.
  \item \emph{Column-sum bound:} For every $j \in [K]$,
  \begin{align}\label{eq:b-bound}
    |b(j)| \estep{a} \bigg|\sum_{i=1}^K \sqrt{p_i r_i}\cdot\sqrt{\frac{p_i}{r_i}}\,k(i,j)\bigg| \lstep{b} \bigg(\sum_{i=1}^K p_i r_i\bigg)^{1/2}\bigg(\sum_{i=1}^K \frac{p_i}{r_i}k(i,j)^2\bigg)^{1/2} \lstep{c} \sqrt{W} ,
  \end{align}
  where (a) writes $p_i = \sqrt{p_i r_i}\sqrt{p_i/r_i}$, (b) is the Cauchy--Schwarz inequality, and (c) uses $W = \sum_{i=1}^K p_i r_i$ and~\eqref{eq:k-2nd}.
\end{enumerate}
We now prove items (a)--(e) in order.

\medskip\noindent\textbf{Proof of (a).}
Using $\kt(i,j) = k(i,j) - \frac{r_i}{W}(b(j) - 1)$, we have
\begin{align*}
  \sum_{j=1}^K p_j \kt(i,j)\, \xb_j &\estep{a} \sum_{j=1}^K p_j k(i,j)\, \xb_j - \frac{r_i}{W}\sum_{j=1}^K p_j \big(b(j) - 1\big) \xb_j \\
  &\estep{b} \xb_i - \frac{r_i}{W}\Bigg(\sum_{l=1}^K p_l \sum_{j=1}^K p_j k(l,j) \xb_j - \sum_{j=1}^K p_j \xb_j\Bigg) \estep{c} \xb_i ,
\end{align*}
where (a) substitutes the definition of $\kt(i,j)$, (b) applies~\eqref{eq:k-reprod} to the first term and expands $b(j) = \sum_{l=1}^K p_l k(l,j)$ in the second, and (c) applies~\eqref{eq:k-reprod} to each $l \in [K]$ to obtain $\sum_{l=1}^K p_l \sum_{j=1}^K p_j k(l,j) \xb_j = \sum_{l=1}^K p_l \xb_l$.

\medskip\noindent\textbf{Proof of (b).}
Summing $\kt(i,j)$ against $p_i$ gives
\begin{align*}
  \sum_{i=1}^K p_i \kt(i,j) = \sum_{i=1}^K p_i k(i,j) - \frac{\sum_{i=1}^K p_i r_i}{W}\big(b(j) - 1\big) = b(j) - \big(b(j) - 1\big) = 1 ,
\end{align*}
where the first equality substitutes the definition of $\kt(i,j)$ and the second uses the definitions $b(j) = \sum_{i=1}^K p_i k(i,j)$ and $W = \sum_{i=1}^K p_i r_i$.

\medskip\noindent\textbf{Proof of (c).}
By the triangle inequality, \eqref{eq:k-mag}, and~\eqref{eq:b-bound},
\begin{align*}
  |\kt(i,j)| \lstep{a} |k(i,j)| + \frac{r_i}{W}\big(|b(j)| + 1\big) \lstep{b} r_i + \frac{r_i}{W}\big(\sqrt W + 1\big) = r_i\big(1 + W^{-1/2} + W^{-1}\big) \lstep{c} 3r_i ,
\end{align*}
where (a) is the triangle inequality, (b) uses $|k(i,j)| \le r_i$ from~\eqref{eq:k-mag} and $|b(j)| \le \sqrt W$ from~\eqref{eq:b-bound}, and (c) uses $W \ge 1$.

\medskip\noindent\textbf{Proof of (d).}
Expanding the square in $\kt(i,j) = k(i,j) - \frac{r_i}{W}(b(j) - 1)$ gives
\begin{align*}
  \sum_{i=1}^K \frac{p_i}{r_i}\kt(i,j)^2
  &\estep{a} \sum_{i=1}^K \frac{p_i}{r_i}k(i,j)^2 - \frac{2(b(j)-1)}{W}\sum_{i=1}^K p_i k(i,j) + \frac{(b(j)-1)^2}{W^2}\sum_{i=1}^K p_i r_i \\
  &\estep{b} \frac{\ell_j^2}{r_j^2} - \frac{2b(j)(b(j)-1)}{W} + \frac{(b(j)-1)^2}{W}
  \estep{c} \frac{\ell_j^2}{r_j^2} - \frac{b(j)^2 - 1}{W} \lstep{d} 1 + \frac1W \lstep{e} 2 ,
\end{align*}
where (a) expands $\kt(i,j)^2$; (b) uses~\eqref{eq:k-2nd}, $\sum_{i=1}^K p_i k(i,j) = b(j)$, and $\sum_{i=1}^K p_i r_i = W$; (c) uses $-2b(b-1) + (b-1)^2 = -(b^2 - 1)$; (d) uses $\ell_j \le r_j$ and $b(j)^2 \ge 0$; and (e) uses $W \ge 1$.

\medskip\noindent\textbf{Proof of (e).}
By definition, $v_i = \frac{1}{r_i}\sum_{j=1}^K p_j \kt(i,j)^2 \ge 0$. Applying items (c) and (d) gives
\begin{align*}
  v_i = \frac{1}{r_i}\sum_{j=1}^K p_j \kt(i,j)^2 \lstep{a} \frac{1}{r_i}\sum_{j=1}^K p_j (3r_i)^2 \estep{b} 9r_i,
  \qquad
  \sum_{i=1}^K p_i v_i \estep{c} \sum_{j=1}^K p_j \sum_{i=1}^K \frac{p_i}{r_i}\kt(i,j)^2 \lstep{d} 2 ,
\end{align*}
where (a) uses $|\kt(i,j)| \le 3r_i$ from item (c), (b) uses $\sum_{j=1}^K p_j = 1$, (c) exchanges the order of summation, and (d) uses item (d) together with $\sum_{j=1}^K p_j = 1$.
\end{proof}

\begin{proof}[Proof of Lemma~\ref{lem:design}]
We verify $\Hb(\brho) \succ 0$ and items (a)--(d) in order.

\medskip\noindent\textbf{Positive definiteness $\Hb(\brho) \succ 0$.}
Since $\{\zb_i\}_{i=1}^K$ spans $\RR^{d+1}$, the uniform distribution gives $\Hb(\mathbf 1/K) \succ 0$. The function $\pb \mapsto \det \Hb(\pb)$ is continuous on the compact set $\Delta_K$, so its maximum is attained and is positive, and any maximizer $\brho^\star$ satisfies $\Hb^\star \coloneqq \Hb(\brho^\star) \succ 0$. By linearity of $\Hb(\cdot)$ and the definition $\brho = \frac12\brho^\star + \frac12(\mathbf 1/K)$,
\begin{align}\label{eq:H-rho-pd}
  \Hb(\brho) = \tfrac12 \Hb^\star + \tfrac12 \Hb(\mathbf 1/K) \succeq \tfrac12 \Hb^\star \succ 0 .
\end{align}

\medskip\noindent\textbf{Proof of (a).}
From $\brho = \frac12 \brho^\star + \frac12(\mathbf 1/K)$ and $\rho^\star_i \ge 0$, we immediately have $\rho_i \ge 1/(2K)$ for all $i \in [K]$.

\medskip\noindent\textbf{Proof of (b).}
Using $\xb_i = \Hb(\brho)^{-1/2} \zb_i$ and $\Hb(\brho) = \sum_{i=1}^K \rho_i \zb_i \zb_i^\top$, we obtain
\begin{align*}
  \sum_{i=1}^K \rho_i \xb_i \xb_i^\top = \Hb(\brho)^{-1/2} \Bigg(\sum_{i=1}^K \rho_i \zb_i \zb_i^\top\Bigg) \Hb(\brho)^{-1/2} = \Hb(\brho)^{-1/2} \Hb(\brho) \Hb(\brho)^{-1/2} = \Ib .
\end{align*}

\medskip\noindent\textbf{Proof of (c).}
For the upper bound $\|\xb_i\|^2 \le 2(d+1)$, note that for every $i \in [K]$ and $\varepsilon \in [0,1]$, the mixture $(1-\varepsilon)\brho^\star + \varepsilon \eb_i$ lies in $\Delta_K$, so first-order optimality of $\brho^\star$ gives
\begin{align*}
  0 \gstep{a} \frac{d}{d\varepsilon}\log\det \Hb\big((1-\varepsilon)\brho^\star + \varepsilon \eb_i\big)\Big|_{\varepsilon = 0^+} \estep{b} \tr\bigl(\Hb^{\star -1}(\zb_i \zb_i^\top - \Hb^\star)\bigr) \estep{c} \zb_i^\top \Hb^{\star -1} \zb_i - (d+1) ,
\end{align*}
where (a) uses that $\brho^\star$ maximizes $\log\det \Hb$ over $\Delta_K$; (b) uses the linearity of $\Hb(\cdot)$, which gives $\Hb((1-\varepsilon)\brho^\star + \varepsilon \eb_i) = \Hb^\star + \varepsilon(\zb_i \zb_i^\top - \Hb^\star)$, and $\frac{d}{d\varepsilon}\log\det(\Hb^\star + \varepsilon \bDelta)|_{\varepsilon=0} = \tr(\Hb^{\star -1}\bDelta)$; and (c) uses $\tr(\Hb^{\star -1}\zb_i \zb_i^\top) = \zb_i^\top \Hb^{\star -1} \zb_i$ and $\tr(\Ib_{d+1}) = d+1$. Combining $\zb_i^\top \Hb^{\star -1} \zb_i \le d+1$ with $\Hb(\brho) \succeq \frac12 \Hb^\star$ from~\eqref{eq:H-rho-pd} (which implies $\Hb(\brho)^{-1} \preceq 2\Hb^{\star -1}$) yields
\begin{align*}
  \|\xb_i\|^2 = \zb_i^\top \Hb(\brho)^{-1} \zb_i \le 2 \zb_i^\top \Hb^{\star -1} \zb_i \le 2(d+1) .
\end{align*}
For the lower bound $\|\xb_i\|^2 \ge 1$, let $\eb = (0,\dots,0,1)^\top \in \RR^{d+1}$ be the last standard basis vector and set $\wb \coloneqq \Hb(\brho)^{1/2} \eb$. Because every $\zb_i = (\ab_i^\top, 1)^\top$ has last coordinate $\eb^\top \zb_i = 1$,
\begin{align*}
  \|\wb\|^2 \estep{a} \eb^\top \Hb(\brho)\, \eb \estep{b} \sum_{i=1}^K \rho_i (\eb^\top \zb_i)^2 \estep{c} 1, \qquad \wb^\top \xb_i \estep{d} \eb^\top \zb_i = 1 ,
\end{align*}
where (a) is the definition of $\wb$, (b) is the definition of $\Hb(\brho)$, (c) uses $\eb^\top \zb_i = 1$ and $\sum_{i=1}^K \rho_i = 1$, and (d) uses $\xb_i = \Hb(\brho)^{-1/2} \zb_i$. The Cauchy--Schwarz inequality then gives $1 = \wb^\top \xb_i \le \|\wb\|\,\|\xb_i\| = \|\xb_i\|$.

\medskip\noindent\textbf{Proof of (d).}
Using $\xb_i = \Hb(\brho)^{-1/2} \zb_i$, the symmetry of $\Hb(\brho)^{-1/2}$, and $\zb_i = (\ab_i^\top, 1)^\top$, we directly compute
\begin{align*}
  \xb_i^\top \Hb(\brho)^{1/2}(\btheta_t^\top, 0)^\top = \zb_i^\top \Hb(\brho)^{-1/2} \Hb(\brho)^{1/2}(\btheta_t^\top, 0)^\top = \zb_i^\top(\btheta_t^\top, 0)^\top = \ab_i^\top\btheta_t .
\end{align*}
\end{proof}

\begin{proof}[Proof of Lemma~\ref{lem:envelope}]
\textbf{Proof of (ii).}
By Lemma~\ref{lem:inner}, $\Qb_t = \Qb(\Qb_{t-1}, \pb_t) \succeq \Qb_{t-1}$, and since the algorithm starts from $\Qb_0 = \Ib$, we have $\Ib = \Qb_0 \preceq \Qb_{t-1} \preceq \Qb_t$ by induction. Because $\Qb \mapsto \xb_i^\top \Qb \xb_i$ is monotone in the Loewner order,
\begin{align*}
  1 \lstep{a} \norm{\xb_i} \estep{b} r_i(\Qb_0) \lstep{c} r_{t-1,i} \lstep{c} r_{ti},
\end{align*}
where (a) is Lemma~\ref{lem:design}(c), (b) uses $\Qb_0 = \Ib$, and (c) uses $\Qb_0 \preceq \Qb_{t-1} \preceq \Qb_t$. This proves item (ii).

\medskip\noindent\textbf{First-order optimality conditions for $\Qb_t$.}
Items (i) and (iii) rest on the first-order optimality conditions of the inner problem $\min_{\Qb \succeq \Qb_{t-1}} J(\Qb, \pb_t)$, which we derive next. Write $\Delta \Qb_t = \Qb_t - \Qb_{t-1}$ and $\Gb_{h,t} = \sum_i \big(h(p_{ti})/r_{ti}\big)\, \xb_i \xb_i^\top$. The function $J(\cdot, \pb_t)$ is differentiable on $\{\Qb \succ 0\}$, with $\nabla_{\Qb} r_i(\Qb) = \xb_i \xb_i^\top / (2 r_i(\Qb))$ and $\nabla_{\Qb} \log\det \Qb = \Qb^{-1}$, so its gradient at $\Qb_t$ is
\begin{align*}
  \nabla J \coloneqq \nabla_{\Qb} J(\Qb_t, \pb_t) = \tfrac12\big(\Gb_{h,t} - \kappa \Qb_t^{-1}\big).
\end{align*}
For every $\Ub \succeq 0$ and $\varepsilon > 0$, the matrix $\Qb_t + \varepsilon \Ub$ is feasible, so the minimality of $\Qb_t$ makes the one-sided derivative $\tr(\nabla J\, \Ub)$ of $\varepsilon \mapsto J(\Qb_t + \varepsilon \Ub, \pb_t)$ at $0$ nonnegative. Taking $\Ub = \vb\vb^\top$ for an arbitrary vector $\vb$ gives $\vb^\top \nabla J\, \vb \ge 0$, that is,
\begin{align}\label{eq:foc-app}
  \Gb_{h,t} \succeq \kappa \Qb_t^{-1} .
\end{align}
In addition, for $\varepsilon \in [0,1]$ the matrix $\Qb_t - \varepsilon\Delta \Qb_t = \Qb_{t-1} + (1-\varepsilon)\Delta \Qb_t$ is feasible, so minimality also gives $\tr(\nabla J\, \Delta \Qb_t) \le 0$. As $\nabla J$ and $\Delta \Qb_t$ are both positive semidefinite, the trace of their product is nonnegative as well, and therefore
\begin{align}\label{eq:cs-app}
  \tr\big((\Gb_{h,t} - \kappa \Qb_t^{-1})\Delta \Qb_t\big) = 0 .
\end{align}

\medskip\noindent\textbf{Proof of (i).}
From~\eqref{eq:foc-app}, we obtain
\begin{align*}
  \kappa \Gb_t \sstep{a} \Gb_{h,t} \sstep{b} \kappa \Qb_t^{-1},
  \qquad\text{hence}\qquad
  \xb_i^\top \Gb_t^{-1} \xb_i \lstep{c} \xb_i^\top \Qb_t\, \xb_i \estep{d} r_{ti}^2,
\end{align*}
where (a) uses $h(p_{ti}) \le \kappa p_{ti}$ (Lemma~\ref{lem:sandwich}) and $\xb_i \xb_i^\top/r_{ti} \succeq 0$; (b) is~\eqref{eq:foc-app}; (c) uses that inversion reverses the Loewner order on positive definite matrices, so $\Gb_t \succeq \Qb_t^{-1}$ implies $\Gb_t^{-1} \preceq \Qb_t$; and (d) is the definition of $r_{ti}$.

\medskip\noindent\textbf{Proof of (iii).}
We use both optimality conditions to limit how fast the largest eigenvalue of $\Qb_t$ can grow. Let $\lambda_1, \dots, \lambda_{d+1}$ denote the eigenvalues of $\Qb_{t-1}^{-1/2} \Qb_t \Qb_{t-1}^{-1/2}$, which are at least $1$ because $\Qb_t \succeq \Qb_{t-1}$. Let $\lambda_t = \lmax(\Qb_t)$ and $s_t = \sqrt{\lambda_t}$, so that $s_0 = 1$, and let $b_T = 4\sqrt2\,(d+1)^{3/2}T$. On the one hand,
\begin{align*}
  \tr(\Gb_{h,t}\Delta \Qb_t) \estep{a} \kappa\tr(\Qb_t^{-1}\Delta \Qb_t) \estep{b} \kappa\tr(\Ib - \Qb_t^{-1}\Qb_{t-1}) \estep{c} \kappa\sum_{k=1}^{d+1} (1 - \lambda_k^{-1}) \lstep{d} \kappa (d+1),
\end{align*}
where (a) is~\eqref{eq:cs-app}; (b) uses $\Delta \Qb_t = \Qb_t - \Qb_{t-1}$; (c) uses that $\Qb_t^{-1}\Qb_{t-1}$ is similar to $(\Qb_{t-1}^{-1/2} \Qb_t \Qb_{t-1}^{-1/2})^{-1}$, whose eigenvalues are $\lambda_k^{-1}$; and (d) uses $\lambda_k \ge 1$. On the other hand,
\begin{align*}
  \tr(\Gb_{h,t}\Delta \Qb_t) \estep{a} \sum_i h(p_{ti}) \frac{\xb_i^\top\Delta \Qb_t \xb_i}{r_{ti}} \gstep{b} \frac{\alpha}{2T\sqrt{2(d+1)\lambda_t}} \sum_i \rho_i\, \xb_i^\top \Delta \Qb_t\, \xb_i \estep{c} \frac{\alpha\,\tr(\Delta \Qb_t)}{2T\sqrt{2(d+1)\lambda_t}} ,
\end{align*}
where (a) is the definition of $\Gb_{h,t}$; (b) uses $h(p_{ti}) \ge \alpha p_{ti} \ge \alpha\rho_i/(2T)$ (Lemma~\ref{lem:sandwich} and $\pb_t \in \Pg$), $r_{ti} \le \norm{\xb_i}\sqrt{\lambda_t} \le \sqrt{2(d+1)\lambda_t}$ (Lemma~\ref{lem:design}(c)) and $\xb_i^\top\Delta \Qb_t \xb_i \ge 0$; and (c) uses $\sum_i \rho_i \xb_i \xb_i^\top = \Ib$ (Lemma~\ref{lem:design}(b)). Combining the two bounds,
\begin{align*}
  \tr(\Delta \Qb_t) \le  \frac{\kappa}{\alpha}\, 2T(d+1)\sqrt{2(d+1)\lambda_t} \le  4\sqrt2\,(d+1)^{3/2} T\, s_t = b_T s_t,
\end{align*}
where the first inequality rearranges the two displays and the second uses $\kappa/\alpha \le 2$ and $s_t = \sqrt{\lambda_t}$. The largest eigenvalue therefore grows slowly:
\begin{align*}
  s_t^2 = \lambda_t \lstep{a} \lambda_{t-1} + \lmax(\Delta \Qb_t) \lstep{b} \lambda_{t-1} + \tr(\Delta \Qb_t) \lstep{c} s_{t-1}^2 + b_T s_t ,
\end{align*}
where (a) is Weyl's inequality (Lemma~\ref{lem:aux-weyl}), (b) uses $\Delta \Qb_t \succeq 0$, and (c) is the previous display. Hence
\begin{align*}
  s_t - s_{t-1} \estep{a} \frac{s_t^2 - s_{t-1}^2}{s_t + s_{t-1}} \lstep{b} \frac{b_T s_t}{s_t + s_{t-1}} \lstep{c} b_T ,
\end{align*}
where (a) multiplies and divides by $s_t + s_{t-1} > 0$, (b) is the previous display, and (c) uses $s_{t-1} \ge 0$. Summing over $t$ and using $s_0 = 1$ gives $s_T \le 1 + T b_T = 1 + 4\sqrt2\,(d+1)^{3/2}T^2$, and
\begin{align*}
  \log\det \Qb_T \lstep{a} (d+1)\log\lambda_T \estep{b} 2(d+1)\log s_T \lstep{c} 2(d+1)\log\big(1 + 4\sqrt2\,(d+1)^{3/2}T^2\big),
\end{align*}
where (a) uses that each of the $d+1$ eigenvalues of $\Qb_T$ is at most $\lambda_T$, (b) uses $\lambda_T = s_T^2$, and (c) uses the bound on $s_T$. This is the first claim of item (iii). The second follows because $\sqrt{2(d+1)} + \tfrac12\log\det \Qb_T \le \sqrt{2(d+1)} + (d+1)\log\big(1 + 4\sqrt2\,(d+1)^{3/2}T^2\big) = A_T$.
\end{proof}

\section{Proofs of the Saddle-Point Lemmas in Appendix~\ref{app:lemmas-B}}\label{app:foundations}

\subsection{The Inner Problem}
\begin{proof}[Proof of Lemma~\ref{lem:inner}]
We first show that $J$ is coercive on $\{\Qb \succeq \Db\}$, which gives existence, then that it is strictly convex along a feasible path between two minimizers, which gives uniqueness, and finally continuity.

Let $\Qb \succeq \Db \succeq \Ib$ and $\lambda = \lmax(\Qb) \ge 1$. For every $j$, $r_j(\Qb) = \sqrt{\xb_j^\top \Qb \xb_j} \le \norm{\xb_j}\sqrt\lambda \le \sqrt{2(d+1)\lambda}$ by Lemma~\ref{lem:design}(c). Therefore
\begin{align*}
  \sum_i \rho_i r_i(\Qb) \gstep{a} \frac{\sum_i \rho_i r_i(\Qb)^2}{\max_j r_j(\Qb)} \estep{b} \frac{\tr\big(\Qb\sum_i \rho_i \xb_i \xb_i^\top\big)}{\max_j r_j(\Qb)} \estep{c} \frac{\tr(\Qb)}{\max_j r_j(\Qb)} \gstep{d} \frac{\lambda}{\sqrt{2(d+1)\lambda}} = \sqrt{\frac{\lambda}{2(d+1)}} ,
\end{align*}
where (a) uses $r_i(\Qb) \le \max_j r_j(\Qb)$; (b) uses $r_i(\Qb)^2 = \tr(\Qb \xb_i \xb_i^\top)$; (c) is Lemma~\ref{lem:design}(b); and (d) uses $\tr(\Qb) \ge \lmax(\Qb) = \lambda$, since $\Qb \succeq 0$, and the bound on $r_j(\Qb)$. Consequently,
\begin{align}\label{eq:coercive}
  J(\Qb,\pb) \ge  \frac{\alpha}{2T}\sum_i \rho_i r_i(\Qb) - \frac{\kappa(d+1)}{2}\log\lambda \ge  \frac{\alpha}{2T}\sqrt{\frac{\lambda}{2(d+1)}} - \frac{\kappa (d+1)}{2}\log\lambda ,
\end{align}
where the first inequality uses $h(p_i) \ge \alpha p_i \ge \alpha\rho_i/(2T)$ (Lemma~\ref{lem:sandwich} and $\pb \in \Pg$), $r_i(\Qb) \ge 0$, and $\log\det \Qb \le (d+1)\log\lambda$, which holds because the $d+1$ eigenvalues of $\Qb$ are at most $\lambda$; and the second is the previous display. At $\Qb = \Db$, on the other hand,
\begin{align*}
  J(\Db,\pb) \lstep{a} \sum_i h(p_i) r_i(\Db) \lstep{b} \kappa \sum_i p_i r_i(\Db) \lstep{c} \kappa\sqrt{2(d+1)\,\lmax(\Db)},
\end{align*}
where (a) uses $\log\det \Db \ge 0$, since $\Db \succeq \Ib$; (b) uses $h(p_i) \le \kappa p_i$ (Lemma~\ref{lem:sandwich}); and (c) uses $r_i(\Db) \le \sqrt{2(d+1)\lmax(\Db)}$ and $\sum_i p_i = 1$. As the right side of~\eqref{eq:coercive} tends to infinity with $\lambda$, every $\Qb \succeq \Db$ with $J(\Qb,\pb) \le J(\Db,\pb)$ satisfies $\lmax(\Qb) \le \bar\lambda$ for some $\bar\lambda$ that depends only on $\lmax(\Db)$ and the constants. We may therefore restrict the minimization to the compact set $\{\Qb \succeq \Db : \lmax(\Qb) \le \bar\lambda\}$, on which $J$ is continuous, so a minimizer exists.

For uniqueness, suppose that $\Qb^{(0)} \ne \Qb^{(1)}$ are both minimizers, and consider the path
\begin{align*}
  \Qb^{(s)} = (\Qb^{(0)})^{1/2} \Mb^s (\Qb^{(0)})^{1/2}, \quad s \in [0,1], \qquad\text{where } \Mb = (\Qb^{(0)})^{-1/2} \Qb^{(1)} (\Qb^{(0)})^{-1/2} .
\end{align*}
In the notation of Lemma~\ref{lem:aux-geomean}, $\Qb^{(s)} = \Qb^{(0)} \#_s \Qb^{(1)}$. Since $\Qb^{(0)}, \Qb^{(1)} \succeq \Db$, the monotonicity of the weighted geometric mean gives $\Qb^{(s)} \succeq \Db \#_s \Db = \Db$, so the path is feasible. Along it, the log-det term is affine in $s$, because $\log\det \Qb^{(s)} = \log\det \Qb^{(0)} + s\log\det \Mb$ and $\log\det \Mb = \log\det \Qb^{(1)} - \log\det \Qb^{(0)}$. The radii, in contrast, are convex along the path. Writing $\Mb = \sum_k \mu_k \ub_k \ub_k^\top$ and setting $\wb_i = (\Qb^{(0)})^{1/2} \xb_i$, $a_{ik} = (\ub_k^\top \wb_i)^2$ and $\ell_k = \log \mu_k$, we have
\begin{align*}
  r_i(\Qb^{(s)})^2 \estep{a} \wb_i^\top \Mb^s \wb_i \estep{b} \sum_k \mu_k^s (\ub_k^\top \wb_i)^2 \estep{c} \sum_k a_{ik} e^{s\ell_k},
  \qquad\text{so}\qquad
  r_i(\Qb^{(s)}) = \exp\Big(\tfrac12 \log \sum_k a_{ik} e^{s\ell_k}\Big),
\end{align*}
where (a) uses the definitions of $\Qb^{(s)}$ and $\wb_i$, (b) uses $\Mb^s = \sum_k \mu_k^s \ub_k \ub_k^\top$, and (c) uses $\mu_k^s = e^{s\ell_k}$. Since log-sum-exp is convex and $\exp$ is increasing and convex, $s \mapsto r_i(\Qb^{(s)})$ is convex. At least one radius is strictly convex. Since $\Mb \ne \Ib$, some $\ell_k$ is nonzero, and since $\{\wb_i\}$ spans $\RR^{d+1}$, some $i$ has $a_{ik} > 0$ for such a $k$. For this $i$, if the values $\{\ell_k : a_{ik} > 0\}$ are not all equal, log-sum-exp is strictly convex, and otherwise it equals a constant plus $s\ell$ with $\ell \ne 0$; in both cases $r_i(\Qb^{(s)})$ is strictly convex. As $h(p_i) > 0$ and the log-det part is affine, the map $s \mapsto J(\Qb^{(s)}, \pb)$ is strictly convex. Hence $J(\Qb^{(1/2)},\pb) < \tfrac12\big(J(\Qb^{(0)},\pb) + J(\Qb^{(1)},\pb)\big) = \min_{\Qb \succeq \Db} J(\Qb,\pb)$, which contradicts minimality, so the minimizer $\Qb(\Db,\pb)$ is unique.

Finally, let $(\Db_n, \pb_n) \to (\Db, \pb)$ and $\Qb_n = \Qb(\Db_n, \pb_n)$. The coercivity bound~\eqref{eq:coercive} and the bound at $\Db_n$ show that $\lmax(\Qb_n)$ is uniformly bounded, so a subsequence converges to some $\bar{\Qb} \succeq \Db$. For any $\Qb' \succeq \Db$, the matrices $\Qb'_n = \Qb' + \norm{\Db_n - \Db}_{\mathrm{op}} \Ib$ satisfy $\Qb'_n \succeq \Db_n$, $\Qb'_n \to \Qb'$ and $J(\Qb_n, \pb_n) \le J(\Qb'_n, \pb_n)$. Since $J$ is jointly continuous on $\{\Qb \succeq \Ib\} \times \Pg$, passing to the limit gives $J(\bar{\Qb}, \pb) \le J(\Qb', \pb)$, and hence $\bar{\Qb} = \Qb(\Db,\pb)$ by uniqueness. Every subsequential limit is the same, so the whole sequence converges.
\end{proof}

\subsection{The Outer Problem}
\begin{proof}[Proof of Lemma~\ref{lem:outer}]
We prove parts (a) and (b) in turn.

\medskip\noindent\textbf{Proof of (a).}
We show that the outer objective is continuous and strongly concave. Let
\begin{align*}
  \Phi_{\Db}(\pb) \coloneqq \min_{\Qb \succeq \Db} J(\Qb,\pb) = J(\Qb(\Db,\pb),\pb) ,
\end{align*}
which is jointly continuous in $(\Db,\pb)$ by Lemma~\ref{lem:inner}. For a fixed $\Qb \succeq \Ib$, the map $\pb \mapsto J(\Qb,\pb) + \norm{\pb}^2/(2\eta)$ is a sum of functions of the individual coordinates, and its second derivative in $p_i$ is
\begin{align*}
  r_i(\Qb)h''(p_i) + \frac1\eta =  -\frac{r_i(\Qb)}{\eta p_i} + \frac1\eta = \frac{1 - r_i(\Qb)/p_i}{\eta} \le  0,
\end{align*}
where the first equality uses $h''(s) = -1/(\eta s)$ and the inequality uses $r_i(\Qb) \ge \norm{\xb_i} \ge 1 \ge p_i$, which holds because $\Qb \succeq \Ib$ and by Lemma~\ref{lem:design}(c). This map is therefore concave. A pointwise minimum of concave functions is concave, so $\Phi_{\Db} + \norm{\cdot}^2/(2\eta)$ is concave as well, that is, $\Phi_{\Db}$ is $1/\eta$-strongly concave. Hence $\pb \mapsto \inner{\Sbb}{\pb} + \Phi_{\Db}(\pb)$ is continuous and strongly concave on the compact convex set $\Pg$ and has a unique maximizer, and by Berge's maximum theorem (Lemma~\ref{lem:aux-berge}), applied on the compact set $\Pg$ to the jointly continuous objective $\inner{\Sbb}{\pb} + \Phi_{\Db}(\pb)$, this maximizer depends continuously on $(\Sbb,\Db)$.

\medskip\noindent\textbf{Proof of (b).}
We show that $\pb^\star$ satisfies the first-order optimality condition of the frozen problem. Write $\cL(\pb, \Qb) = \inner{\Sbb}{\pb} + J(\Qb,\pb)$ and $\tilde\Phi(\pb) = \inner{\Sbb}{\pb} + \Phi_{\Db}(\pb)$, take any $\qb \in \Pg$, and for $\varepsilon \in (0,1]$ let $\pb_\varepsilon = \pb^\star + \varepsilon(\qb - \pb^\star)$, which lies in $\Pg$ by convexity, and $\Qb_\varepsilon = \Qb(\Db, \pb_\varepsilon)$. Then
\begin{align*}
  \cL(\pb_\varepsilon, \Qb_\varepsilon) \estep{a} \tilde\Phi(\pb_\varepsilon) \lstep{b} \tilde\Phi(\pb^\star) \estep{c} \cL(\pb^\star, \Qb^\star) \lstep{d} \cL(\pb^\star, \Qb_\varepsilon),
\end{align*}
where (a) and (c) use that $\Qb_\varepsilon$ and $\Qb^\star$ minimize $J(\cdot, \pb_\varepsilon)$ and $J(\cdot, \pb^\star)$ over $\{\Qb \succeq \Db\}$; (b) uses that $\pb^\star$ maximizes $\tilde\Phi$ over $\Pg$ (part (a)); and (d) uses that $\Qb^\star$ minimizes $J(\cdot, \pb^\star)$ and $\Qb_\varepsilon \succeq \Db$ is feasible. Hence $\cL(\pb_\varepsilon, \Qb_\varepsilon) - \cL(\pb^\star, \Qb_\varepsilon) \le 0$. By the mean value theorem, this difference equals $\varepsilon \inner{\nabla_{\pb} \cL(\bar{\pb}_\varepsilon, \Qb_\varepsilon)}{\qb - \pb^\star}$ for some $\bar{\pb}_\varepsilon$ on the segment between $\pb^\star$ and $\pb_\varepsilon$. We divide by $\varepsilon$ and let $\varepsilon \to 0$. Since $\Qb_\varepsilon \to \Qb^\star$ by Lemma~\ref{lem:inner} and $\partial_{p_i}\cL = S_i + h'(p_i) r_i(\Qb)$ is jointly continuous, we obtain
\begin{align*}
  \inner{\nabla_{\pb} \cL(\pb^\star, \Qb^\star)}{\qb - \pb^\star} \le 0 \qquad \text{for all } \qb \in \Pg .
\end{align*}
The map $\pb \mapsto \cL(\pb, \Qb^\star)$ is strictly concave, since $h$ is strictly concave and $r_i(\Qb^\star) > 0$, and the last display is its first-order optimality condition on $\Pg$; so $\pb^\star$ is its unique maximizer. Finally, $\cL(\pb, \Qb^\star) = \inner{\Sbb}{\pb} + \sum_i r_i(\Qb^\star) h(p_i) - \tfrac\kappa2 \log\det \Qb^\star$, and the last term does not depend on $\pb$, which proves part (b).
\end{proof}

\section{Auxiliary Lemmas}\label{app:aux}

\begin{lemma}[{Azuma--Hoeffding Inequality~\citep{hoeffding1963probability,azuma1967weighted}}]\label{lem:aux-azuma}
Let $(M_t)_{t \in [T]}$ be a martingale difference sequence with respect to a filtration $(\cF_t)$, with $|M_t| \le 1$ for every $t$. Then for every $\delta' \in (0,1)$, with probability at least $1 - \delta'$,
\begin{align*}
  \sum_{t=1}^T M_t \le \sqrt{2T\log(1/\delta')} .
\end{align*}
\end{lemma}

\begin{lemma}[{Weyl's Inequality~\citep[Theorem~4.3.1]{horn2012matrix}}]\label{lem:aux-weyl}
For symmetric matrices $\Ab$ and $\Bb$ of the same size, $\lmax(\Ab + \Bb) \le \lmax(\Ab) + \lmax(\Bb)$.
\end{lemma}

\begin{lemma}[{Berge's Maximum Theorem~\citep[see][Theorem~17.31]{berge1963topological,aliprantis2006infinite}}]\label{lem:aux-berge}
Let $X$ be a compact subset of $\RR^K$, let $\Theta$ be a metric space, and let $f : X \times \Theta \to \RR$ be continuous. If, for every $\theta \in \Theta$, the function $f(\cdot, \theta)$ has a unique maximizer $\xb^\star(\theta)$ over $X$, then $\theta \mapsto \xb^\star(\theta)$ is continuous.
\end{lemma}

\begin{lemma}[{Monotonicity of the Weighted Geometric Mean~\citep[see][Chapter~4]{kubo1980means,bhatia2007positive}}]\label{lem:aux-geomean}
For positive definite matrices $\Ab$ and $\Bb$ and $s \in [0,1]$, let $\Ab \#_s \Bb = \Ab^{1/2}\big(\Ab^{-1/2} \Bb \Ab^{-1/2}\big)^s \Ab^{1/2}$. Then $\Ab \#_s \Ab = \Ab$, and $\Ab \preceq \Ab'$, $\Bb \preceq \Bb'$ imply $\Ab \#_s \Bb \preceq \Ab' \#_s \Bb'$.
\end{lemma}

\bibliography{reference}

@article{auer2002nonstochastic,
  title={The nonstochastic multiarmed bandit problem},
  author={Auer, Peter and Cesa-Bianchi, Nicol{\`o} and Freund, Yoav and Schapire, Robert E},
  journal={SIAM Journal on Computing},
  volume={32},
  number={1},
  pages={48--77},
  year={2002},
  publisher={SIAM}
}

@inproceedings{beygelzimer2011contextual,
  title={Contextual bandit algorithms with supervised learning guarantees},
  author={Beygelzimer, Alina and Langford, John and Li, Lihong and Reyzin, Lev and Schapire, Robert},
  booktitle={Proceedings of the Fourteenth International Conference on Artificial Intelligence and Statistics (AISTATS)},
  pages={19--26},
  year={2011}
}

@inproceedings{bogunovic2021stochastic,
  title     = {Stochastic Linear Bandits Robust to Adversarial Attacks},
  author    = {Ilija Bogunovic and Arpan Losalka and Andreas Krause and Jonathan Scarlett},
  booktitle = {Proceedings of the 24th International Conference on Artificial Intelligence and Statistics (AISTATS)},
  series    = {Proceedings of Machine Learning Research},
  volume    = {130},
  pages     = {991--999},
  year      = {2021}
}

@inproceedings{bubeck2012towards,
  title     = {Towards Minimax Policies for Online Linear Optimization with Bandit Feedback},
  author    = {Bubeck, S{\'e}bastien and Cesa-Bianchi, Nicol{\`o} and Kakade, Sham M.},
  booktitle = {Proceedings of the 25th Annual Conference on Learning Theory (COLT)},
  series    = {Proceedings of Machine Learning Research},
  volume    = {23},
  pages     = {41.1--41.14},
  year      = {2012},
  publisher = {PMLR}
}

@inproceedings{gupta2019better,
  title     = {Better Algorithms for Stochastic Bandits with Adversarial Corruptions},
  author    = {Anupam Gupta and Tomer Koren and Kunal Talwar},
  booktitle = {Proceedings of the 32nd Conference on Learning Theory (COLT)},
  series    = {Proceedings of Machine Learning Research},
  volume    = {99},
  pages     = {1562--1578},
  year      = {2019}
}

@inproceedings{he2022nearly,
  title     = {Nearly Optimal Algorithms for Linear Contextual Bandits with Adversarial Corruptions},
  author    = {Jiafan He and Dongruo Zhou and Tong Zhang and Quanquan Gu},
  booktitle = {Advances in Neural Information Processing Systems (NeurIPS)},
  volume    = {35},
  pages     = {34614--34625},
  year      = {2022}
}

@book{lattimore2020bandit,
  title     = {Bandit Algorithms},
  author    = {Lattimore, Tor and Szepesv{\'a}ri, Csaba},
  publisher = {Cambridge University Press},
  year      = {2020},
  isbn      = {9781108486828},
  doi       = {10.1017/9781108571401}
}

@inproceedings{lykouris2018stochastic,
  title     = {Stochastic Bandits Robust to Adversarial Corruptions},
  author    = {Thodoris Lykouris and Vahab Mirrokni and Renato Paes Leme},
  booktitle = {Proceedings of the 50th Annual ACM SIGACT Symposium on Theory of Computing (STOC)},
  publisher = {ACM},
  pages     = {114--122},
  year      = {2018},
  doi       = {10.1145/3188745.3188918}
}

@article{kubo1980means,
  title={Means of positive linear operators},
  author={Kubo, Fumio and Ando, Tsuyoshi},
  journal={Mathematische Annalen},
  volume={246},
  number={3},
  pages={205--224},
  year={1980},
  publisher={Springer}
}

@book{bhatia2007positive,
  title={Positive Definite Matrices},
  author={Bhatia, Rajendra},
  year={2007},
  publisher={Princeton University Press}
}

@article{azuma1967weighted,
  title={Weighted sums of certain dependent random variables},
  author={Azuma, Kazuoki},
  journal={Tohoku Mathematical Journal},
  volume={19},
  number={3},
  pages={357--367},
  year={1967}
}

@article{hoeffding1963probability,
  title={Probability inequalities for sums of bounded random variables},
  author={Hoeffding, Wassily},
  journal={Journal of the American Statistical Association},
  volume={58},
  number={301},
  pages={13--30},
  year={1963}
}

@book{horn2012matrix,
  title={Matrix Analysis},
  author={Horn, Roger A and Johnson, Charles R},
  edition={2nd},
  year={2012},
  publisher={Cambridge University Press}
}

@book{berge1963topological,
  title={Topological Spaces: Including a Treatment of Multi-Valued Functions, Vector Spaces, and Convexity},
  author={Berge, Claude},
  year={1963},
  publisher={Macmillan}
}

@book{aliprantis2006infinite,
  title={Infinite Dimensional Analysis: A Hitchhiker's Guide},
  author={Aliprantis, Charalambos D and Border, Kim C},
  edition={3rd},
  year={2006},
  publisher={Springer}
}

@inproceedings{awerbuch2004adaptive,
  title     = {Adaptive routing with end-to-end feedback: distributed learning and geometric approaches},
  author    = {Awerbuch, Baruch and Kleinberg, Robert D.},
  booktitle = {Proceedings of the 36th Annual ACM Symposium on Theory of Computing (STOC)},
  pages     = {45--53},
  year      = {2004},
  publisher = {ACM},
  doi       = {10.1145/1007352.1007367}
}

@inproceedings{mcmahan2004online,
  title     = {Online Geometric Optimization in the Bandit Setting Against an Adaptive Adversary},
  author    = {McMahan, H. Brendan and Blum, Avrim},
  booktitle = {Proceedings of the 17th Annual Conference on Learning Theory (COLT)},
  series    = {Lecture Notes in Computer Science},
  pages     = {109--123},
  year      = {2004},
  publisher = {Springer},
  doi       = {10.1007/978-3-540-27819-1_8}
}

@inproceedings{dani2007price,
  title     = {The Price of Bandit Information for Online Optimization},
  author    = {Dani, Varsha and Hayes, Thomas P. and Kakade, Sham M.},
  booktitle = {Advances in Neural Information Processing Systems (NIPS)},
  volume    = {20},
  year      = {2007},
  publisher = {Curran Associates, Inc.}
}

@inproceedings{abernethy2008competing,
  title     = {Competing in the Dark: An Efficient Algorithm for Bandit Linear Optimization},
  author    = {Abernethy, Jacob and Hazan, Elad and Rakhlin, Alexander},
  booktitle = {Proceedings of the 21st Annual Conference on Learning Theory (COLT)},
  pages     = {263--274},
  year      = {2008}
}

@inproceedings{bartlett2008high,
  title     = {High-Probability Regret Bounds for Bandit Online Linear Optimization},
  author    = {Bartlett, Peter L. and Dani, Varsha and Hayes, Thomas P. and Kakade, Sham M. and Rakhlin, Alexander and Tewari, Ambuj},
  booktitle = {Proceedings of the 21st Annual Conference on Learning Theory (COLT)},
  pages     = {335--342},
  year      = {2008}
}

@article{cesa2012combinatorial,
  title     = {Combinatorial bandits},
  author    = {Cesa-Bianchi, Nicol{\`o} and Lugosi, G{\'a}bor},
  journal   = {Journal of Computer and System Sciences},
  volume    = {78},
  number    = {5},
  pages     = {1404--1422},
  year      = {2012},
  publisher = {Elsevier},
  doi       = {10.1016/j.jcss.2012.01.001}
}

@inproceedings{lee2020bias,
  title     = {Bias no more: high-probability data-dependent regret bounds for adversarial bandits and {MDPs}},
  author    = {Lee, Chung-Wei and Luo, Haipeng and Wei, Chen-Yu and Zhang, Mengxiao},
  booktitle = {Advances in Neural Information Processing Systems (NeurIPS)},
  volume    = {33},
  pages     = {15522--15533},
  year      = {2020},
  publisher = {Curran Associates, Inc.}
}

@inproceedings{zimmert2022return,
  title     = {Return of the bias: Almost minimax optimal high probability bounds for adversarial linear bandits},
  author    = {Zimmert, Julian and Lattimore, Tor},
  booktitle = {Proceedings of the 35th Conference on Learning Theory (COLT)},
  series    = {Proceedings of Machine Learning Research},
  volume    = {178},
  pages     = {3285--3312},
  year      = {2022},
  publisher = {PMLR}
}

@inproceedings{foster2020adapting,
  title     = {Adapting to Misspecification in Contextual Bandits},
  author    = {Foster, Dylan J. and Gentile, Claudio and Mohri, Mehryar and Zimmert, Julian},
  booktitle = {Advances in Neural Information Processing Systems (NeurIPS)},
  volume    = {33},
  pages     = {11478--11489},
  year      = {2020},
  publisher = {Curran Associates, Inc.}
}

@inproceedings{neu2020efficient,
  title     = {Efficient and robust algorithms for adversarial linear contextual bandits},
  author    = {Neu, Gergely and Olkhovskaya, Julia},
  booktitle = {Proceedings of the 33rd Conference on Learning Theory (COLT)},
  series    = {Proceedings of Machine Learning Research},
  volume    = {125},
  pages     = {3049--3068},
  year      = {2020},
  publisher = {PMLR}
}

@article{kiefer1960equivalence,
  title     = {The Equivalence of Two Extremum Problems},
  author    = {Kiefer, J. and Wolfowitz, J.},
  journal   = {Canadian Journal of Mathematics},
  volume    = {12},
  pages     = {363--366},
  year      = {1960},
  publisher = {Canadian Mathematical Society},
  doi       = {10.4153/CJM-1960-030-4}
}

@article{zimmert2021tsallis,
  title     = {{T}sallis-{INF}: An Optimal Algorithm for Stochastic and Adversarial Bandits},
  author    = {Zimmert, Julian and Seldin, Yevgeny},
  journal   = {Journal of Machine Learning Research},
  volume    = {22},
  number    = {28},
  pages     = {1--49},
  year      = {2021}
}

@inproceedings{lee2021achieving,
  title     = {Achieving Near Instance-Optimality and Minimax-Optimality in Stochastic and Adversarial Linear Bandits Simultaneously},
  author    = {Lee, Chung-Wei and Luo, Haipeng and Wei, Chen-Yu and Zhang, Mengxiao and Zhang, Xiaojin},
  booktitle = {Proceedings of the 38th International Conference on Machine Learning (ICML)},
  series    = {Proceedings of Machine Learning Research},
  volume    = {139},
  pages     = {6142--6151},
  year      = {2021},
  publisher = {PMLR}
}

@inproceedings{kuroki2024best,
  title     = {Best-of-Both-Worlds Algorithms for Linear Contextual Bandits},
  author    = {Kuroki, Yuko and Rumi, Alberto and Tsuchiya, Taira and Vitale, Fabio and Cesa-Bianchi, Nicol{\`o}},
  booktitle = {Proceedings of the 27th International Conference on Artificial Intelligence and Statistics (AISTATS)},
  series    = {Proceedings of Machine Learning Research},
  volume    = {238},
  pages     = {1216--1224},
  year      = {2024},
  publisher = {PMLR}
}

@article{li2019stochastic,
  title   = {Stochastic Linear Optimization with Adversarial Corruption},
  author  = {Yingkai Li and Edmund Y. Lou and Liren Shan},
  journal = {arXiv preprint arXiv:1909.02109},
  year    = {2019}
}

@inproceedings{bogunovic2020corruption,
  title     = {Corruption-Tolerant {G}aussian Process Bandit Optimization},
  author    = {Ilija Bogunovic and Andreas Krause and Jonathan Scarlett},
  booktitle = {Proceedings of the 23rd International Conference on Artificial Intelligence and Statistics (AISTATS)},
  series    = {Proceedings of Machine Learning Research},
  volume    = {108},
  pages     = {1071--1081},
  year      = {2020}
}

@inproceedings{lattimore2020learning,
  title     = {Learning with Good Feature Representations in Bandits and in {RL} with a Generative Model},
  author    = {Tor Lattimore and Csaba Szepesv{\'a}ri and Gellert Weisz},
  booktitle = {Proceedings of the 37th International Conference on Machine Learning (ICML)},
  series    = {Proceedings of Machine Learning Research},
  volume    = {119},
  pages     = {5662--5670},
  year      = {2020}
}

@article{zhao2021linear,
  title   = {Linear Contextual Bandits with Adversarial Corruptions},
  author  = {Heyang Zhao and Dongruo Zhou and Quanquan Gu},
  journal = {arXiv preprint arXiv:2110.12615},
  year    = {2021}
}

@inproceedings{ding2022robust,
  title     = {Robust Stochastic Linear Contextual Bandits Under Adversarial Attacks},
  author    = {Qin Ding and Cho-Jui Hsieh and James Sharpnack},
  booktitle = {Proceedings of the 25th International Conference on Artificial Intelligence and Statistics (AISTATS)},
  series    = {Proceedings of Machine Learning Research},
  volume    = {151},
  pages     = {7111--7123},
  year      = {2022}
}

@inproceedings{wei2022model,
  title     = {A Model Selection Approach for Corruption Robust Reinforcement Learning},
  author    = {Chen-Yu Wei and Christoph Dann and Julian Zimmert},
  booktitle = {Proceedings of the 33rd International Conference on Algorithmic Learning Theory (ALT)},
  series    = {Proceedings of Machine Learning Research},
  volume    = {167},
  pages     = {1043--1096},
  year      = {2022}
}

@inproceedings{liu2024corruption,
  title     = {Corruption-Robust Linear Bandits: Minimax Optimality and Gap-Dependent Misspecification},
  author    = {Haolin Liu and Artin Tajdini and Andrew Wagenmaker and Chen-Yu Wei},
  booktitle = {Advances in Neural Information Processing Systems (NeurIPS)},
  volume    = {37},
  pages     = {24277--24325},
  year      = {2024}
}

@inproceedings{ye2023corruption,
  title     = {Corruption-Robust Algorithms with Uncertainty Weighting for Nonlinear Contextual Bandits and {M}arkov Decision Processes},
  author    = {Chenlu Ye and Wei Xiong and Quanquan Gu and Tong Zhang},
  booktitle = {Proceedings of the 40th International Conference on Machine Learning (ICML)},
  series    = {Proceedings of Machine Learning Research},
  volume    = {202},
  pages     = {39834--39863},
  year      = {2023}
}

@inproceedings{yu2025corruption,
  title     = {Corruption-Robust Variance-aware Algorithms for Generalized Linear Bandits under Heavy-tailed Rewards},
  author    = {Qingyuan Yu and Euijin Baek and Xiang Li and Qiang Sun},
  booktitle = {Proceedings of the 41st Conference on Uncertainty in Artificial Intelligence (UAI)},
  series    = {Proceedings of Machine Learning Research},
  volume    = {286},
  pages     = {4826--4843},
  year      = {2025}
}

@article{cheng2026adversarial,
  title   = {Adversarial Bandit Optimization with Globally Bounded Perturbations to Linear Losses},
  author  = {Zhuoyu Cheng and Kohei Hatano and Eiji Takimoto},
  journal = {arXiv preprint arXiv:2603.26066},
  year    = {2026}
}

@inproceedings{lykouris2021corruption,
  title     = {Corruption-Robust Exploration in Episodic Reinforcement Learning},
  author    = {Thodoris Lykouris and Max Simchowitz and Alex Slivkins and Wen Sun},
  booktitle = {Proceedings of the 34th Conference on Learning Theory (COLT)},
  series    = {Proceedings of Machine Learning Research},
  volume    = {134},
  pages     = {3242--3245},
  year      = {2021}
}

@inproceedings{jun2018adversarial,
  title     = {Adversarial Attacks on Stochastic Bandits},
  author    = {Kwang-Sung Jun and Lihong Li and Yuzhe Ma and Xiaojin Zhu},
  booktitle = {Advances in Neural Information Processing Systems (NeurIPS)},
  volume    = {31},
  year      = {2018}
}

@inproceedings{liu2019data,
  title     = {Data Poisoning Attacks on Stochastic Bandits},
  author    = {Fang Liu and Ness Shroff},
  booktitle = {Proceedings of the 36th International Conference on Machine Learning (ICML)},
  series    = {Proceedings of Machine Learning Research},
  volume    = {97},
  pages     = {4042--4050},
  year      = {2019}
}

@inproceedings{garcelon2020adversarial,
  title     = {Adversarial Attacks on Linear Contextual Bandits},
  author    = {Evrard Garcelon and Baptiste Rozi{\`e}re and Laurent Meunier and Jean Tarbouriech and Olivier Teytaud and Alessandro Lazaric and Matteo Pirotta},
  booktitle = {Advances in Neural Information Processing Systems (NeurIPS)},
  volume    = {33},
  pages     = {14362--14373},
  year      = {2020}
}

@inproceedings{di2025nearly,
  title     = {Nearly Optimal Algorithms for Contextual Dueling Bandits from Adversarial Feedback},
  author    = {Qiwei Di and Jiafan He and Quanquan Gu},
  booktitle = {Proceedings of the 42nd International Conference on Machine Learning (ICML)},
  series    = {Proceedings of Machine Learning Research},
  volume    = {267},
  pages     = {13513--13545},
  year      = {2025}
}

@inproceedings{yang2020adversarial,
  title     = {Adversarial Bandits with Corruptions},
  author    = {Lin Yang and Mohammad Hajiesmaili and Mohammad Sadegh Talebi and John C. S. Lui and Wing Shing Wong},
  booktitle = {Advances in Neural Information Processing Systems (NeurIPS)},
  volume    = {33},
  pages     = {20994--21005},
  year      = {2020}
}

@article{lewis1978finite,
  title     = {Finite dimensional subspaces of {$L_p$}},
  author    = {Lewis, D. R.},
  journal   = {Studia Mathematica},
  volume    = {63},
  number    = {2},
  pages     = {207--212},
  year      = {1978}
}

@inproceedings{cohen2015lp,
  title     = {{$L_p$} row sampling by {L}ewis weights},
  author    = {Cohen, Michael B. and Peng, Richard},
  booktitle = {Proceedings of the 47th Annual ACM Symposium on Theory of Computing (STOC)},
  pages     = {183--192},
  year      = {2015}
}
\bibliographystyle{ims}
\end{document}